\documentclass[sigconf]{acmart}

\AtBeginDocument{%
	\providecommand\BibTeX{{%
			\normalfont B\kern-0.5em{\scshape i\kern-0.25em b}\kern-0.8em\TeX}}}

\acmConference[]{}{}{}
\acmPrice{}
\acmISBN{}
\usepackage{paralist}
\usepackage{multirow}
\usepackage{subcaption}
\usepackage{xcolor}
\usepackage{tikz}
\usepackage{amsmath,amsfonts}
\usepackage{algorithmic}
\usepackage{textcomp}
\usepackage{url}
\usepackage{balance}

\usepackage{multicol}
\usepackage{makecell}
\usepackage{arydshln}
\usepackage[bottom]{footmisc}
\usepackage{comment}

\newtheorem{definition}{Definition}
\newtheorem{prob}{Problem}

\usepackage[most]{tcolorbox}
\usetikzlibrary{shadows}

\newcommand{\candidateset}{C}
\newcommand{\resolution}{M}
\newcommand{\classifier}{\mu}
\newcommand{\problemName}{{\sf MIER}}
\newcommand{\problemNameSpace}{{\sf MIER} }
\newcommand{\modelName}{{\sf FlexER}}
\newcommand{\modelNameSpace}{{\sf FlexER} }
\newcommand{\mier}{multiple intents entity resolution}
\newcommand{\er}{entity resolution }
\newcommand{\ernospace}{entity resolution}
\newcommand{\Er}{Entity resolution }
\newcommand{\ER}{Entity Resolution }
\newcommand{\numB}{two }

\newcommand{\update}[1]{\textcolor{black}{{#1}}} % Update
\newcommand*{\TechReport}{}

\begin{document}
\newpage

\title{FlexER: Flexible Entity Resolution for Multiple Intents (Technical Report)}
%\subtitle{One Size does not Always Fit All}

\author{Bar Genossar}
\affiliation{%
%	\institution{Technion -- Israel Institute of Technology}
	\institution{Technion}
	\city{Haifa}
	\country{Israel}}
\email{sbargen@campus.technion.ac.il}

\author{Roee Shraga}
\affiliation{%
	\institution{Northeastern University}
	\city{Boston}
	\country{MA, USA}}
\email{r.shraga@northeastern.edu}

\author{Avigdor Gal}
\affiliation{%
%	\institution{Technion -- Israel Institute of Technology}
	\institution{Technion}
	\city{Haifa}
	\country{Israel}}
\email{avigal@technion.ac.il}

\begin{abstract}
Entity resolution, %(ER), 
a longstanding problem of data cleaning and integration, aims at identifying data records that represent the same real-world entity. Existing approaches treat \er as a universal task, assuming the existence of a {\bf single} interpretation of a real-world entity and focusing only on finding matched records, separating corresponding from non-corresponding ones, with respect to this single interpretation. However, in real-world scenarios, where \er is part of a more general data project, downstream applications may have varying interpretations of real-world entities 
%only require resolution of records that refer to the same entity but may also seek to match records that share different levels of commonality, 
relating, for example, to various user needs.
%granularity levels of the resolution. 
In what follows, we introduce the problem of {\em \mier} ({\sf MIER}), %\bg{We might change it to a new name}, 
an extension to the universal (single intent) \er task. As a solution, we propose {\sf FlexER}, utilizing contemporary solutions to universal \er tasks to solve \mier. %\bg{The following sentence is too long and clumsy. Let's change it}
\update{{\sf FlexER} %\bg{We might change it as well} 
addresses the problem as a multi-label classification problem. It combines intent-based representations of tuple pairs using a multiplex graph representation that serves as an input to a graph neural network (GNN). {\sf FlexER} learns intent representations and improves the outcome to multiple resolution problems.} A large-scale empirical evaluation introduces a new benchmark and, using also \numB well-known benchmarks, shows that {\sf FlexER} effectively solves the {\sf MIER} problem and outperforms the state-of-the-art for a universal \ernospace.
\end{abstract}

\maketitle

\section{Introduction}
\label{sec:intro}

An essential component of any data science lifecycle involves data preparation. Accordingly, contemporary data projects intensively invest in large-scale data cleaning and integration techniques to combine data from multiple heterogeneous sources into meaningful and valuable information. \update{Consider, for example, a data-intensive organization ({\em e.g.}, an online shopping company) that continuously collects data on user queries to better understand their needs. The (mostly implicit) intents of users may vary. While some users are interested in basketball shoes, regardless of their brand, others may be more interested in buying branded products, caring less whether shoes are designated as running shoes.} %Such a company may need to integrate user query data from multiple sources and add to it history of purchases.}%\bg{It feels like there is a missing sentence here. Maybe something like: "Users might search for the desired product with a different level of granularity and specification, as a function of their knowledge, desires, and perception of their requirements."}

At the heart of the data preparation realm lies the \emph{\er}%Entity Resolution} (ER) 
task (with variations of entity matching, record linkage, deduplication, and more), which has been extensively studied over the past decades (see books and surveys~\cite{elmagarmid2006duplicate,getoor2012entity,christen2012data}). \Er is a data integration task that aims at identifying same real-world entities 
%\bg{find another term instead of entities} \ag{I try to create a flexible description of an entity}
that are represented by different data records (instances). Specifically, \er can be used to ``clean'' a dataset from duplicate tuples referring to the same entity, offering a useful post processing tool for integrating multiple data sources. %\update{Also, in a setting where data arrives in a streaming fashion, online duplicate detection is useful to ensure data is connected and aggregated in a correct fashion.}\bg{What is the contribution of the last sentence? None of the reviewers said mentioned online ER, and this is not part of our work.}

\Er techniques for resolving duplicates in a dataset were developed over years of research. Traditional methods focused on string similarity~\cite{levenshtein1966binary,jaro1989advances,lin1998information,jaro1995probabilistic} and rule-based methods~\cite{singla2006entity,singh2017synthesizing}. Learning-based approaches were also suggested~\cite{bilenko2003adaptive,konda2016magellan}, followed by deep learning methods in recent years~\cite{joty2018distributed,mudgal2018deep,fu2019end,kasai2019low,zhao2019auto,fu2020hierarchical,li2020grapher}. Following a common practice for text processing, the use of \emph{pre-trained language models}, and specifically, BERT-based models~\cite{devlin:2018bert}, was also introduced for \er~\cite{li2020deep,brunner2020entity,li2021improving,peeters2021dual}.

The aforementioned methods share a common assumption regarding the universality of the \er problem, assuming the existence of a {\bf single} interpretation to the notion of a real-world entity.
% and focusing only on finding matched records, separating corresponding from non-corresponding ones. a standalone task independent of the more general data integration/cleaning (or even data science) problem they aim to solve. 
In particular, most solutions aim to create a single clean view of a dataset, by separating corresponding from non-corresponding record pairs, with respect to this single interpretation. 

The universal property of \er solutions is being challenged by the development of applications that require the ability to provide a fine-grained (or personalized) analysis, offering services that are tailored to specific user needs and requirements~\cite{yoganarasimhan2019search,boratto2021countering}. \update{For example, online shops aim at personalizing shopping experience to the needs of individual users, having to infer such needs from mainly implicit feedback~\cite{kelly2003implicit}, such as clicks and mouse movements.} In such a setting, having a universal resolution facility is likely to cripple the resolution process, failing to cater to the needs of all applications at all times. To illustrate the need for a non-universal (flexible) resolution, we use the \emph{AmazonMI} dataset, a newly suggested benchmark based on amazon products data (see Section~\ref{sec:benchmarks}), in the following motivating example.

\subsection{Motivating Example: From Record Duplication to \ER Interpretation}\label{sec:motivation}

\begin{figure}[t]
	\centering
	\includegraphics[width=0.68\columnwidth]{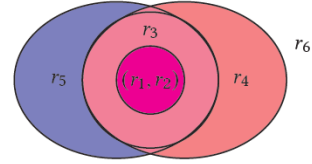}
	\caption{Possible \er solutions for different entity interpretations over the tuples from Table~\ref{tab:example}.}
	\label{fig:example}
\end{figure}

Table~\ref{tab:example} depicts an excerpt of the AmazonMI dataset to illustrate the differences between the universal approach to \er and the flexible \er we present and address in this work. Figure~\ref{fig:example} offers visual illustration to the multiplicity of entity interpretations in the example.

Record duplication is usually the result of discordant representations (\emph{e.g.,} multi-lingual, synonyms, capitalizations), changes in the data over time, typos, \emph{etc.} For example, records $r_1$ and $r_2$ in Table~\ref{tab:example} refer to the same %real-world entity -- a 
pair of Nike basketball shoes called ``Men's Lunar Force 1 Duckboot.'' The difference between $r_1$ and $r_2$ originates from capitalization issues (Nike vs. NIKE) and additional specification (\emph{e.g.,} color). Contrarily, $r_1$ and $r_6$ will most likely be conceived as %are obviously %refer to completely 
different, as the latter refers to a book. %entities. 

The pair ($r_1,r_3$) is an example of a pair that, under a certain interpretation, does not refer to the same entity, representing two different Nike basketball shoe variants. Yet, under a different interpretation, both $r_1$ and $r_3$ are Nike basketbsall shoes. Similarly, $r_1$ differs from $r_4$ and $r_5$ since they represent a different type of Nike shoes (basketball vs. running) and basketball shoes of different brands (Nike vs. Adidas), respectively, but may be determined to be representatives of the same entity (Nike shoes and basketball shoes, respectively) under a broader interpretation. \update{Such different interpretation may be attributed to different user tastes or different contexts. For example, a pro basketball player would be more sensitive to differences between basketball shoes than a user that seeks shoes for a neighborhood afternoon fun game.}%\bg{Like my previous suggestion, we can say that user' interpretation can be the results of his knowledge or awareness to its needs. For example, if I would like to buy a basketball shoe, my query will probably be more specific than a query of parent who looks for new basketball show for his child and is not familiar with the differences between the models. If this is the case, matching between different shoes which still fall under the same broader definition (e.g., NIKE basketball shoes) can be beneficial in satisfying the user's need for information}

Over the years, \er solutions resolved a single entity interpretation, %overcome the aforementioned challenges and 
detecting %sole (exact) duplicate 
$r_1$ and $r_2$ as duplicates, to create a (single) clean dataset, \emph{e.g.,} by only preserving $r_1$. \update{The multiple intent phenomenon challenges any downstream application that requires entity matching and involves personalization to user’s needs. Beyond online shopping, multiple intents can be frequently found in the domain of recommendation systems, where user's implicit feedback ({\em e.g.}, item selection and search queries) can serve in understanding a user’s intent for better responding to her needs~\cite{kelly2003implicit}. }

\begin{comment}
\begin{table}[t]
	\caption{Amazon Product Dataset Sample ($D_{ex}$)}
	\label{tab:example}
	\scalebox{0.775}{
		\begin{tabular}{lll}
%			\toprule
			\hline
			\textbf{$r_{id}$} & \textbf{Product title} & \textbf{Category(s)} \\
			\hline
%			\midrule
			$r_1$ &  Nike Men's Lunar Force 1 & Clothing, Shoes \& Jewelry, \\
			& Duckboot & Men, Shoes, Athletic, \\
			&& Team Sports, Basketball\\\hline
			$r_2$ &  NIKE Men Lunar Force 1 & Clothing, Shoes \& Jewelry, \\
			&  Duckboot, Black/Dark & Men, Shoes, Athletic \\
			& Loden-BROGHT Crimson & Team Sports, Basketball \\\hline
			$r_3$ &  NIKE Men's Air Max Stutter & Clothing, Shoes \& Jewelry, \\
			& Step Ankle-High Basketball Shoe & Men, Shoes, Athletic, \\
			& & Team Sports, Basketball\\\hline
			$r_4$ &  Nike Men's Air Max & Clothing, Shoes \& Jewelry, \\
			& 2016 Running Shoe & Men, Shoes, Athletic,\\
			&& Running, Road Running\\\hline
			$r_5$ &  adidas Performance Men's & Clothing, Shoes \& Jewelry, \\
			&  D Rose 6 Boost Primeknit Basketball & Men, Shoes, Athletic, \\
			&&Team Sports, Basketball\\\hline
			$r_6$ & The Man Who Tried to Get Away & Books, Mystery, \\
			&&Thriller \& Suspense\\
%			\bottomrule
			\hline
	\end{tabular}
}
\end{table}
\end{comment}
\begin{table}[t]
	\caption{Amazon Product Dataset Sample ($D_{ex}$)}
	\label{tab:example}
	\scalebox{0.775}{
		\begin{tabular}{ll}
			%			\toprule
			\hline
			\textbf{$r_{id}$} & \textbf{Product title}\\
			\hline
			%			\midrule
			$r_1$ &  Nike Men's Lunar Force 1 Duckboot\\\hline
			$r_2$ &  NIKE Men Lunar Force 1 Duckboot,\\
			& Black/Dark Loden-BROGHT Crimson\\\hline
			$r_3$ &  NIKE Men's Air Max Stutter\\
			& Step Ankle-High Basketball Shoe\\\hline
			$r_4$ &  Nike Men's Air Max 2016 Running Shoe\\\hline
			$r_5$ &  adidas Performance Men's\\
			&  D Rose 6 Boost Primeknit Basketball\\\hline
			$r_6$ & The Man Who Tried to Get Away\\
			%			\bottomrule
			\hline
		\end{tabular}
	}
\end{table}

%\def\maincircle{(0,0) circle (.5cm)}
%\def\rthreecircle{(0,0) circle (1cm)}
%\def\rfourcircle{(.5,0) ellipse (1.5cm and 1.15cm)}
%\def\rfifthcircle{(-.5,0) ellipse (1.5cm and 1.15cm)}

%\begin{figure}[t]
%	\begin{center}
%	\begin{tikzpicture}[scale=0.6, every node/.style={transform shape}]
%	% The last trick is to cheat and use transparency
%	\begin{scope}[blend group = soft light]
%	\fill[magenta] \maincircle;
%%	\fill[green] \rthreecircle;
%	\fill[red!50!white] \rfourcircle;
%	\fill[blue!50!white] \rfifthcircle;
%	\end{scope}
%	\draw \maincircle node [] {$(r_1, r_2)$};
%	\draw \rthreecircle node [above=.5cm] {$r_3$};
%	\draw \rfourcircle node [right=.6cm] {$r_4$};
%	\draw \rfifthcircle node [left=.6cm] {$r_5$};
%	\draw  node [above=.5cm, right=2cm] {$r_6$};
%	\end{tikzpicture}
%	\caption{Possible \er solutions for different entity interpretations over the tuples from Table~\ref{tab:example}}
%	\label{fig:example}
%	\end{center}
%\end{figure}

\subsection{Main Contributions}

The motivating example offers an intuitive description of a scenario where a single dataset may serve as a basis for multiple clean views to be generated by an \er solution, suggesting different interpretations of an entity in each such view. We use the term {\em intent} (formally defined in Section~\ref{sec:MI}) to reflect user preferences that, in the scope of this paper, relate to the %granularity level 
interpretation of the output view. We are, in particular, interested in intents that are unknown apriori and therefore cannot be constructed using data that exists in the database. Rather, such intents are known only through the training set that labels tuple pairs to be matched under a given interpretation. % \ag{The following is extremely important to assist in convincing the reader. I didn't nail it yet so if you have any bright ideas let me know:} \bg{I like it. In addition, I wonder if we can also add something about how machines can interpret intents (it can't, but it can learn patterns and inter-relationships between them) vs how human can. What do you think?} 
A typical such scenario \update{can be found in recommendation systems where feedback is collected from users, either implicitly or explicitly, to be used to indicate intent. Another possible scenario may be motivated using machine learning algorithm for entity matching, where} %In particular, such feedback may be collected using 
different tuning parameters yield different outcomes even for the same algorithm. \update{For example, see the use-case of Yad VaShem, the Jewish Holocaust museum in Jerusalem~\cite{DBLP:conf/sigmod/SagiGBBA16}}.

In this work, we define the problem of {\em multiple intents \ernospace} (\problemName), an extension to the universal (single intent) \er task. The \problemNameSpace task considers multiple intents when creating a solution for potential downstream applications that involve \er problems. In the absence of human interpretation to intents \update{(recall the use of implicit feedback in recommendation systems to indicate intent)}, we aim at training a model to offer such interpretation based on training data and intent cross learning. We propose \modelName, a solution to the \problemNameSpace problem that utilizes contemporary solutions to universal \er tasks (\emph{e.g.,} DITTO~\cite{li2020deep}). %to solve \problemName. 
\modelNameSpace positions the problem as a multi-label classification problem and combines intent-based representations of tuple pairs by creating an expressive multiplex graph. Graph neural network (GNN) uses the multiplex graph to learn latent relationships among intents, which in turn improves the outcome to multiple resolution problems. Our empirical evaluation uses \numB well-known benchmarks and proposes a new dataset for \problemName, showing that \modelNameSpace provides accurate results for \problemName. Moreover, \modelNameSpace outperforms state-of-the-art for the universal \er problem. Specifically, the paper offers the following four contributions.  

%We propose \modelNameSpace as a solution to the \problemNameSpace problem, utilizing contemporary solutions to universal ER tasks (\emph{e.g.,} DITTO~\cite{li2020deep}) to solve parallel single intent ER. \modelNameSpace combines the independent intent solutions using a graph convolutional network (GCN) to improve outcome to multiple resolution problems. Our empirical evaluation uses three well-known benchmarks and proposes a new dataset for \problemName, showing that our proposed solution provides accurate results for \problemName. Moreover, \modelNameSpace outperforms state-of-the-art for the universal ER problem. Specifically, the paper offers the following four contributions:  
\begin{compactenum}
	\item A formulation of a new variation of the \er problem, \problemName, %properly representing 
	addressing the challenge of integrating and cleaning data in a multi-intent environment (Section~\ref{sec:mier}).
	\item A solution for \problemName, \modelName, that
		\begin{compactenum}
			\item employs a novel representation of intent interrelationships using a multiplex graph (Section~\ref{sec:Gcreate}); and
			\item enriches the input vector-based representation by training a graph neural network, to create multiple intents-aware prediction (Section~\ref{sec:Gmessage}).
%			\item trains graph neural networks to identify multiple intents, as part of the prediction phase of \er (Section~\ref{sec:Gmessage}); and
%			\item enriches the input vector-based representation using message passing, rather than directly use it for prediction (Section~\ref{sec:Gmessage} as well).
		\end{compactenum}	
	\item A large-scale empirical evaluation showing the effectiveness of \modelNameSpace in solving \problemName, while also outperforming the state-of-the-art on standard, universal \er (Section~\ref{sec:eval}). 
	\item An open source access to \modelNameSpace implementation and the newly suggested benchmark dataset~\footnote{\url{https://github.com/BarGenossar/FlexER/}}.%\footnote{\url{https://github.com/BarGenossar/FlexER/}}
\end{compactenum}

\noindent We provide the building blocks of our model in Section~\ref{sec:model}. In addition, Section~\ref{sec:methodology} offers some baselines to be compared against \modelName. A discussion of related work (Section~\ref{sec:related}) and final remarks (Section~\ref{sec:con}) conclude the paper.

\section{Model and Problem Definiton}
\label{sec:model}

We present next an entity resolution %(ER) 
model (Section~\ref{sec:ER}) and introduce resolution intent (Section~\ref{sec:MI}) as an extension to the universal \er model to address multiple intents (Section~\ref{sec:mier}).
\ifdefined\TechReport
Table~\ref{tab:symbols} summarizes the notations used throughout the following two sections.
\begin{table}[t]
	\begin{center}
		\scalebox{.85}{\begin{tabular}{cc}
				\midrule
				\textbf{Notation} & \textbf{Meaning} \\\midrule
				$D$  & Set of data records \\
				$E$ & Set of entities \\
				$\theta$ & Mapping from $D$ to $E$ \\
				$C$ & Set of candidate record pairs \\
				$(r_i, r_j)$ & A candidate pair in C \\
				$M$ & Resolution \\ 
				$\pi$ & Intent \\
				$\Pi$ & Set of intents \\
				$y^{\pi}_{ij}$ & The true label of $(r_i, r_j)$ with respect to intent $\pi$ \\
				$\classifier_{\pi}(r_i, r_j)$ & The matcher prediction of $(r_i, r_j)$ with respect to intent $\pi$ \\
				\hline
		\end{tabular}}
		\caption{Notation Table}
		\label{tab:symbols}
	\end{center}
\end{table}
\else

\fi

\subsection{(Single Intent) Entity Resolution}
\label{sec:ER}

\Er has several (analogous) definitions in the literature. We now present the model definition we use in this paper.%In this work we follow the view of \er as a clustering problem~\cite{cohen2002learning,papadakis2020three}, which is typically associated with a broader data cleaning procedure. %An illustration of the ER process is given in Figure~\ref{fig:er}. 

Let $D=\{r_1,r_2,...,r_n\}$ be a set of data records (\emph{dataset}) and $E=\{e_1,e_2,...,e_m\}$ a set of real-world entities ($m \leq n$). Each 
record is associated with an entity in $E$ using an {\em entity mapping} (mapping for short) $\theta: D\rightarrow E$.  %The goal of ER is to generate a set of \emph{real-world entities} $E=\{e_1,e_2,...,e_m\}$ ($m\leq n$) representing a (single) clean view of the dataset. 
Whenever $\theta$ is unknown, for example, due to the absence of unique keys to identify entities, \er solutions aim to pair records in $D$ such that if $\{r_i,r_j\}\subseteq D$ are paired together then $\theta(r_i)=\theta(r_j)$. $D$ is usually characterized by a set of attributes $A=\{a_1,a_2,...,a_k\}$, such that a record $r_i= \langle  r_i.a_1,r_i.a_2,...,r_i.a_k \rangle$ is assigned with values to all attributes (some of which may be null values). It is worth noting that in a data integration scenario where multiple data sources are involved, schema matching~\cite{shraga2020adnev} can provide an integrated attribute set and $D$ can be composed as a union of the sources. 

\begin{figure}[t]
	\centering
	\includegraphics[width=\columnwidth]{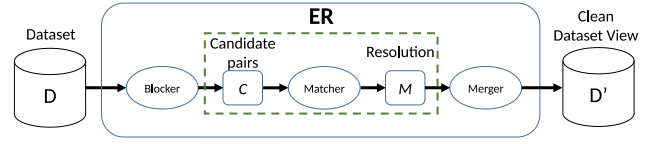}
	\caption{A (Single Intent) Entity Resolution Process. The matching phase in a dashed square is the focus of this work.}
	\label{fig:er}
\end{figure}

\Er is typically a three phase problem (see Figure~\ref{fig:er}). It starts with a blocking phase~\cite{christen2012data}, which aims to reduce the number of comparisons between records by eliminating record pairs $(r_i,r_j)\in D\times D$ for which $\theta(r_i)\not=\theta(r_j)$. Such pairs can be deduced from the construction of $D$. For example, in the case of a clean-clean resolution, where the integration task aims at integrating two clean data sources, each source is assumed to be duplicate-free by itself. Thus, two records that belong to the same data source cannot be matched together. \Er solutions in the literature use the blocking phase as a tool to improve performance, by applying heuristics to assess the chance of $\theta(r_i)=\theta(r_j)$ and eliminating pairs that are unlikely to match. The blocking phase generates a set of \emph{candidate record pairs} $\candidateset\subseteq D\times D$, over which \emph{matchers} perform pairwise record pair comparisons. 

During the matching phase, which is the focus of our work, a matcher assigns \emph{likelihood} (similarity) scores to record pairs $(r_{i}, r_{j})\in \candidateset$ that endured the blocking phase. The likelihood score can be viewed as an estimation for the probability that $\theta(r_i)=\theta(r_j)$.
Applying a threshold over the likelihood scores yields $\resolution\subseteq\candidateset$, a {\em resolution}, containing record pairs that the matcher resolve to represent the same real-world entity. The relationship between resolution and entity mapping can be defined as follows.
%$(r_{i}, r_{j})\in M$ for which $\theta(r_i)=\theta(r_j)$.
%to determine for each pair $(r_{i}, r_{j})\in M$ whether  $\theta(r_i)=\theta(r_j)$. %the records $r_{i}$ and $r_{j}$ refer to the same real-world entity. over $D$ is a set of corresponding record pairs. 

\begin{definition}[Resolution Satisfaction]\label{def:intentSat}
	Let $D$ be a dataset, $E$ an entity set, $\resolution \subseteq\candidateset\subseteq D\times D$, and $\theta:D\rightarrow E$ an entity mapping. $\resolution$ {\em satisfies} $\theta$ (denoted $\resolution\models\theta$) if $\forall \left(r_i,r_j \right)\in \candidateset$,  $\left(r_i,r_j \right)\in \resolution \Leftrightarrow\theta(r_i)=\theta(r_j)$. 
\end{definition}

Finally, induced by the pairs in $\resolution$, the merging phase involves deriving $D^{\prime}$, a clean view of $D$, by choosing equivalence class representatives~\cite{elmagarmid2006duplicate}  
(assuming reflexivity, symmetry, and transitivity). %$\theta$. To provide a clean view of the dataset, we can merge records that refer to the same entity by, for example, selecting a representative record. the set $E$, can be devised by combining the matched record pairs in $R$ using a transitivity assumption, \emph{i.e.,} if $(r_{i},r_{j})\in R$ and $(r_{j},r_{k})\in R$ then also $(r_{i},r_{k})$ refer to the same entity. For completeness, we note that we assume that any chosen mapping that generates $E$ from $D$ creates an equivalence relation, thus ensuring symmetric and reflexivity as well. 

\begin{sloppypar}
\begin{example}
	Recall Table~\ref{tab:example} (termed $D_{ex}$) and let $\candidateset_{ex} = D_{ex}\times D_{ex}$. Assume that some matcher assigns likelihood scores of $0.9$ to $(r_1,r_2)$, $0.8$ to $(r_1,r_3)$ and a likelihood score lower than $0.5$ to all other record pairs in $\candidateset_{ex}$. Applying a threshold of $0.5$, we obtain a resolution of $\resolution_{ex} = \{(r_1,r_2), (r_1,r_3)\}$, clustered into $\{\{r_{1}, r_{2}, r_3\},\{r_4\},\{r_5\},\{r_6\}\}$ with a possible clean view $D^{\prime}=\{r_1,r_4,r_5,r_6\}$.   
\end{example}
\end{sloppypar}

%ER matchers use similarity as indication for equivalence. Contemporary ER solutions use learning-based matchers and typically cast the problem as \emph{binary classification} for which the record pairs in a set $\candidateset$ are classified as matched (equivalent) ($1$) or non-matched ($0$). Record pair representation is the core ingredient of learning-based matchers. Prior art used multiple similarity scores~\cite{konda2016magellan}, while recent works use record information (attribute values) to obtain a latent feature representation~\cite{joty2018distributed,mudgal2018deep}. Learning-based matchers assume the availability of a (labeled) training set $\mathcal{D}_{train} = \{(r_{i}, r_{j}),y_{ij}\}_{(r_i,r_j)\in \candidateset_{train}}$ to train a matcher to distinguish matching from non-matching records. $\candidateset_{train}$ denotes a training set of record pairs and $y_{ij} = 1$ if  $\theta(r_i)=\theta(r_j)$ and $0$ otherwise. 

%Contemporary ER solutions utilize deep learning to obtain a latent representation of $(r_i,r_j)$, which can be a combination of the individual representations of $r_i$ and $r_j$~\cite{joty2018distributed,zhao2019auto,li2020grapher}, attribute-based~\cite{mudgal2018deep,fu2019end,fu2020hierarchical}, or combined~\cite{li2020deep,brunner2020entity}. A unique characteristic of the deep learning-based matchers is that the record pair representation can be fine-tuned according to a specific task. This becomes especially beneficial in our setting as we aim to create a record pair representation tuned for each intent. 

\Er matchers use similarity as a proxy to equivalence. Contemporary \er solutions use learning-based matchers and typically cast the problem as a binary classification problem, classifying record pairs in a set $\candidateset$ as matched (equivalent) ($1$) or non-matched ($0$). Record pair representation is the core ingredient of learning-based matchers. Prior art used multiple similarity scores~\cite{konda2016magellan}, while recent works use record information (attribute values) to obtain a latent feature representation, based on individual representations~\cite{joty2018distributed,zhao2019auto,li2020grapher}, attribute representations~\cite{mudgal2018deep,fu2019end,fu2020hierarchical,cappuzzo2020creating}, or by creating a combined representation~\cite{li2020deep,brunner2020entity}. Learning-based matchers typically assume the availability of a (labeled) training set $\mathcal{D}_{train} = \{(r_{i}, r_{j}),y_{ij}\}_{(r_i,r_j)\in \candidateset_{train}}$ to train a matcher, where %to distinguish matching from non-matching records.
 $\candidateset_{train}$ denotes a record pair training set and $y_{ij} = 1$ if  $\theta(r_i)=\theta(r_j)$ and $0$ otherwise. 

\begin{example}[DITTO Matcher]\label{ex:ditto}
	DITTO~\cite{li2020deep} is an example of a deep learning-based state-of-the-art matcher %that currently provides state-of-the-art results 
	for the \er task. DITTO matches entities over candidate record pairs $\candidateset\subseteq D\times D^{\prime}$ from two data sources $D$ and $D^{\prime}$. \update{DITTO serializes and tokenizes record pairs, adding a special token (termed $[cls]$) to support classification (see~\cite{devlin:2018bert} for details). Then, it applies fine-tuning of a pre-trained transformer-based language (BERT-based) model %\footnote{See~\cite{devlin:2018bert} for additional details on BERT-based models.} 
	to obtain latent representations (dimension of 768) for each tuple pair. These latent representations are used for binary classification using a linear layer.} To enrich learning, DITTO injects domain knowledge, augments training data, and summarizes long strings. 
\end{example}

\subsection{Resolution Intents for Multiple Entity Interpretations}
\label{sec:MI}

Standard resolution methods provide an adequate solution for universal \ernospace, a standalone task with a single interpretation (which we refer to as the \emph{equivalence intent}). Yet, as motivated in Section~\ref{sec:motivation}, downstream data cleaning/integration applications may require different interpretations for the same input, to be formalized next as \emph{multiple resolution intents}. \update{We start with defining an intent.}
%\vskip -0.15in
\update{
\begin{definition}[Resolution Intent]\label{def:resolution_intent}
	Let $D$ be a dataset, $E$ an entity set and $\theta$ mapping from $D$ to $E$. A {\em resolution intent} (intent for short) is a pair $(E, \theta)$. %such that $E$ is set of entities against which $\theta$ is computed.
\end{definition} 
}
%In this work we offer an extension to the universal view of \ernospace, pointedly reflected in the use of a single mapping from records to real-world entities, to include multiple intents supporting different interpretations. 
To better understand the meaning of multiple intents, we note that the universal view of \er implicitly assumes a single entity set $E$ by which the \er solution must abide. Such an entity set is not explicitly known, yet typically referred to abstractly as a ``real-world entity." %a complimentary justification to the ones suggested in the literature. 
We argue that an entity set of choice may vary according to user needs, and that different users may seek different %solutions 
interpretations (and accordingly different solutions) for the same dataset. For illustration purposes, recall the motivating example (Section~\ref{sec:motivation}) demonstrating a case where users may seek varying entity interpretations. An attempt to seek a universal solution whenever multiple interpretations exist reduces the quality of the resolution outcome. For example, an intent to resolve $r_1$ and $r_3$ as matching records would fail under the universal \er solution, which solely dictates $\theta(r_1)=\theta(r_2)$.%$r_1$ with $r_2$. 

%We term the %resolution of a 
%universal \er as having an \emph{equivalence} intent, assuming the existence of a single entity set $E$ (see Section~\ref{sec:ER}) against which $\theta(r_i)=\theta(r_j)$ is evaluated for records $r_i$ and $r_j$. %assuming that the broader data science task that requires \er is solely interested in equivalent record pairs. 
We argue that the universal \er has an underlying \emph{equivalence} intent and assumes the existence of a single entity set $E$ and mapping $\theta$ such that if $\theta(r_i)=\theta(r_j)$, then the records $r_i$ and $r_j$ refer to the same real world entity in $E$ (see Section~\ref{sec:ER}).
\Er matchers, such as DITTO (see Example~\ref{ex:ditto}), 
all aim at solving the \er problem with an equivalence intent. The framework that we suggest here takes into account resolution intent in the \er process. 

%A {\em resolution intent} (intent for short) is an entity set $E$ against which a mapping $\theta$ is computed. Such a mapping, while can be theoretically defined exhaustively, is pragmatically unknown. For ease of representation, we label intents using %and can be defined symbolically by using
A mapping $\theta$, while can be theoretically defined exhaustively, is pragmatically unknown. For ease of representation, we label intents using %and can be defined symbolically by using 
predicates such as \emph{``same category''} or \emph{``same brand''} when offering illustrating examples. Such labeling is for illustration purposes only and does not indicate that these predicates are known or can be derived from the data at hand. Rather, the model ``perceives'' the intents as sets of inputs (as explained in Example~\ref{ex:ditto}) and corresponding labels, and learns the underlying relationships between intents (to be  explained in Section~\ref{sec:flexer}). We use $\pi$ to denote an %such a symbolic 
intent. %It is worth noting that in our framework, the model is agnostic to intents characteristics, hence utilizing domain knowledge for inferring transitivity is impossible. Instead,  We further note that detecting such intents is beyond the scope of this work.

%Recall that records represent compound objects (such as products in Table~\ref{tab:example}), sharing multiple levels of commonalities. For example, records $r_1$ and $r_4$ in Table~\ref{tab:example} share the same brand Nike. In this work, we focus on intents that alternate along granularity levels and use the concept of intent to model varying resolution needs. 

\begin{example}\label{ex:intents}
	Recall Table~\ref{tab:example} with the dataset $D_{ex}$ and let $\pi_{eq}$ be an equivalence intent. We obtain $\resolution_{eq}=\{(r_1,r_2)\}$ under $\pi_{eq}$, using the subscript to relate the resolution with the intent description. 

	Let $\pi_{brand}$ be an intent of \emph{``same brand.''} Then, we obtain that all pairs $i,j\in \{1,2,3,4\}$ are part of a resolution under $\pi_{brand}$. This means that the user having a $\pi_{brand}$ intent, intends to resolve the records $\{r_1,r_2,r_3,r_4\}$ (Nike products). When looking at a \emph{``same category''} intent $\pi_{cat.}$, the resolution becomes less obvious. For example, $r_1$, $r_2$, $r_3$, and $r_5$ share the same exact category (jointly representing basketball shoes). However, if we zoom out and look at the shoes category, a resolution satisfying the intent should also include $r_4$. Such differences are reflected in the definition of intents, taking into account the overlap between categories. An additional intent $\pi_{brand+cat.}$ may combine the two, referring to \emph{``same brand''} and \emph{``same category,''} to produce a resolution with record pairs $i,j\in \{1,2,3\}$ (Nike basketball shoes). It is worth noting here that all intents are determined using the same input dataset. 
\end{example}

\subsection{Multiple Intents Entity Resolution (\protect\problemName)}
\label{sec:mier}

Equipped with an intent definition as an entity set $E$ against which a mapping $\theta$ is computed, we next define the problem of \emph{multiple intents entity resolution} (\problemName). 
A \problemNameSpace involves a set of (possibly related) entity mappings for a set of intents $\mathcal{E} = \{E_{1}, E_{2}, \cdots, E_{P}\}$, offering multiple interpretations to resolve the entities in $D$, each serving as a solution for a respective intent. 
%A one-size-fits-all resolution provides an adequate solution for universal ER, a standalone task with a single equivalence intent. Yet, some data cleaning/integration challenges (recall Section~\ref{sec:motivation}) may involve multiple intents. Therefore, instead of performing a universal ER, we argue for enhancing ER to support multiple outcomes for multiple intents. A \problemNameSpace involves a set of (possibly related) entity mappings for a set of intents $\mathcal{E} = \{E_{1}, E_{2}, \cdots, E_{P}\}$, offering multiple ways to divide $D$, each serving as a solution for a respective intent. 
%The \problemNameSpace problem is defined as follows. 

%\begin{prob}[Multiple Intents Entity Resolution]\label{def:PD}
%	Let $D$ be a dataset, $\candidateset\subseteq D\times D$, $\mathcal{E} = \{E_{1}, E_{2}, \cdots, E_{P}\}$ a set of intents, and $\Theta = \{\theta_{1}, \theta_{2}, \cdots, \theta_{P}\}$ a set of mappings from $D$ to the respective intent. A \emph{multiple intents entity resolution} (\problemName) seeks a set of resolutions $\mathcal{\resolution} = \{\resolution_{1}, \resolution_{2}, \cdots, \resolution_{P}\}$ over $\candidateset$ such that for each $1\leq p\leq P, \resolution_p\models\theta_p$.
%\end{prob}
\begin{prob}[Multiple Intents Entity Resolution]\label{def:PD}
	Let $D$ be a dataset, $\candidateset\subseteq D\times D$ and $\{(E_{1},\theta_{1}) \cdots, (E_{P},\theta_{P})\}$ a set of intents.. A \emph{multiple intents entity resolution} (\problemName) seeks a set of resolutions $\mathcal{\resolution} = \{\resolution_{1}, \cdots, \resolution_{P}\}$ over $\candidateset$ such that for each $1\leq p\leq P, \resolution_p\models\theta_p$.
\end{prob}
%Intuitively, a solution for \problemNameSpace should provide multiple solutions, each constituting a clean dataset \emph{view} of $D$ for a different intent.
Intuitively, a solution for \problemNameSpace should provide multiple solutions, each constituting a clean dataset \emph{view} of $D$ for a different intent.
\begin{sloppypar}
%\addtocounter{example}{-1}
%\begin{example}[cont]
\begin{example}
	Recalling again our running example, now associated with a set of intents $\{\pi_{eq}, \pi_{brand}, \pi_{cat.}, \pi_{brand+cat.}\}$, a possible \problemNameSpace solution over $D_{ex}$ returns the resolutions $\{(r_1, r_2)\}$, $\{(r_1, r_2), (r_2, r_3), (r_3, r_4)\}$, $\{(r_1, r_2), (r_2, r_3), (r_3, r_5)\}$, $\{(r_1, r_2), (r_2, r_3)\}$ for the intents $\pi_{eq}$, $\pi_{brand}$, $\pi_{cat.}$, and $\pi_{brand+cat.}$, respectively (illustrated in Figure~\ref{fig:example}). Out of these resolutions, we can generate the following clean views of $D_{ex}$, $\{r_1, r_3, r_4, r_5, r_6\}$, $\{r_1, r_5, r_6\}$, $\{r_1, r_4, r_6\}$, $\{r_1, r_4, r_5, r_6\}$ (heuristically choosing representatives by order).
\end{example}
\end{sloppypar}

\subsection{Intents Interrelationships}
\label{sec:intentrelations}

As illustrated in Example~\ref{ex:intents}, intents are not simply standalone entity sets.
Interrelationships among intents may provide useful hints on how to mix and match resolution results aiming to satisfy different intents. We begin with defining \emph{overlapping intents}.

\begin{definition}[Overlapping Intents]\label{def:overlap}
	Let $D$ be a dataset, $\candidateset\subseteq D\times D$, $(E, \theta)$ and $(E^{\prime}, \theta^{\prime})$ two intents, where $E$ and $E^{\prime}$ are entity sets and $\theta$ and $\theta^{\prime}$ mappings from $D$ to $E$ and $E^{\prime}$, respectively. $\resolution, \resolution^{\prime}$ are resolutions such that $\resolution\models\theta$ and $\resolution^{\prime}\models\theta^{\prime}$. $(E, \theta)$ and $(E^{\prime}, \theta^{\prime})$ \emph{overlap} if $\exists (r_i, r_j)\in \candidateset: (r_i, r_j)\in \resolution\land (r_i, r_j)\in \resolution^{\prime}$.
\end{definition} 

%\begin{definition}[Overlapping Intents]\label{def:overlap}
%	Let $D$ be a dataset, $\candidateset\subseteq D\times D$, $E$ and $E^{\prime}$ two intents, $\theta$ and $\theta^{\prime}$ mappings from $D$ to $E$ and $E^{\prime}$, respectively, and $\resolution, \resolution^{\prime}$ resolutions such that $\resolution\models\theta$ and $\resolution^{\prime}\models\theta^{\prime}$. $E$ and $E^{\prime}$ \emph{overlap} if $\exists (r_i, r_j)\in \candidateset: (r_i, r_j)\in \resolution\land (r_i, r_j)\in \resolution^{\prime}$.
%\end{definition} 

%\addtocounter{example}{-1}
%\begin{example}[cont]
	For example, intents $\pi_{eq}$ and $\pi_{brand}$ overlap since the record pair $(r_1,r_2)$ is part of both resolutions.
%\end{example}

In Section~\ref{sec:flexer} we present our methodology to train a model to identify interrelationships among intents using multiplex graphs and a GNN. At times, it may be possible to identify special cases of overlapping intents. For example, consider \emph{subsumed intents}, defined as follows.

\begin{definition}[Subsumed Intents]\label{def:subsum}
	Let $D$ be a dataset, $\candidateset\subseteq D\times D$, $(E, \theta)$ and $(E^{\prime}, \theta^{\prime})$ two intents, and $\resolution, \resolution^{\prime}$ resolutions such that $\resolution\models\theta$ and $\resolution^{\prime}\models\theta^{\prime}$. $(E^{\prime}, \theta^{\prime})$ is a \emph{sub-intent} of $(E, \theta)$ if $\nexists (r_i, r_j)\in \candidateset: (r_i, r_j)\not\in \resolution\land (r_i, r_j)\in \resolution^{\prime}$.
\end{definition}
%\addtocounter{example}{-1}
%\begin{example}[cont]
	For example, $\pi_{eq}$ is a sub-intent of $\pi_{brand}$. $\pi_{brand}$ and $\pi_{cat.}$ are overlapping but not subsumed intents since, for 
	example, $(r_1, r_5)$ is in $\resolution_{cat.}$ but not in $\resolution_{brand}$.
%\end{example}

Subsumed intents may be the result of algorithm parameter tuning in a way that weakens the matching criteria so that the conditions of Definition~\ref{def:subsum} are met.

The resolution process can benefit from intents interrelationships. For example, having a resolution for an equivalence intent, we already know that all resolved record pairs with this intent will also be part of a resolution for a \emph{``same brand''} intent, the former subsumed by the latter. 

Whereas intents can be the outcome of inherent characteristics of the data (\emph{e.g., } ontology, available attributes, {\em etc.}), they can also be formed according to user specifications. For intuition sake,
%\begin{example}\label{ex:intentrelations}
	recall Table~\ref{tab:example} and assume the availability of sales data. The R\&D department wishes to examine the impact a product's store location has on its selling (\emph{e.g.,} how placing products of the same category but different brands in neighboring shelves contributes to the selling of the lower-price brand). In parallel, the purchasing department wants to determine the desired brand quantities to acquire. %Given a set of candidate pairs alongside intents-based matching judgment, 
	Although intents meaning by themselves are not necessarily unknown, interrelationships among them can be derived, either deterministically or stochastically. In this case, the R\&D department focuses on category-level matching, while the purchasing department on brand-level matching. By receiving a set of candidate pairs with corresponding binary matching decisions of these intents, a machine learning model can derive, despite being agnostic to the intents' underlying essence, that a match with respect to the latter intent implies a match with respect to the former. %In the opposite direction, if a candidate pair does not match according to the first intent it probably does not match according to the second one. %Anyhow, interrelationships between intents might not always be deterministic; Therefore, a system capable of interpreting them will likely to provide an enriched inference of each intent as standalone task. 
%\end{example}

%\section{Addressing Multiple Intents Entity Resolution Problem}
\section{Flexible Entity Resolution}
\label{sec:methodology}

%\bg{Any suggestion of a better name to this section?}
As a preface to presenting our proposed solution, we first describe a general approach for addressing flexible \ernospace, \er with more than one intent (Section~\ref{sec:baselines}). Then, we provide two baseline solutions to \problemName. The first, termed {\em in-parallel}, treats multiple intents as a set of independent single intent problems. In this setting, each intent is considered individually and provides an independent solution for a single intent \er (Section~\ref{sec:multi}). An alternative approach, solving all intents jointly using multi-label learning, is presented in Section~\ref{sec:prior}. 

\subsection{\ER Beyond a Single Intent}\label{sec:baselines}
Just like other machine learning methods, learning-based matchers (\emph{e.g.,} DITTO, see Example~\ref{ex:ditto}), are trained (or fine-tuned) over a given set of training examples. In the context of universal single intent \er (Section~\ref{sec:ER}), a set of record pairs $\candidateset_{train}$ is labeled with respect to an equivalence intent (Section~\ref{sec:MI}) to form a training set $\mathcal{D}_{train} = \{(r_{i}, r_{j}),y_{ij}\}_{(r_i,r_j)\in \candidateset_{train}}$. Specifically, a record pair $(r_{i}, r_{j})\in \candidateset_{train}$ is labeled $y_{ij}=1$ if the record pair $(r_i,r_j)$ satisfies $\theta_{eq}(r_i)=\theta_{eq}(r_j)$. 

%\begin{sloppypar}
%\rs{do we actually need the term ``resolution creation mapping''? Can't we just refer to the classifier as a matcher?}
Recalling that an intent induces a boolean space of record pairs, an \er problem can be cast as a binary classification problem. Accordingly, a matcher\footnote{We use the term {\em matcher} rather than {\em classifier}, noting that in the case of supervised learning these terms can be used interchangeably.} %Then, given $\mathcal{D}_{train}(\pi_{p})$, we can train a matcher $\classifier_{p}$ using Eq.~\ref{eq:ce}.} 
aims to learn a resolution creation mapping $\classifier_{eq}:\candidateset\rightarrow \{0,1\}$, which separates record pairs that correspond (%$\theta_{eq}(r_i)=\theta_{eq}(r_j)$, 
$\classifier_{eq}(r_i, r_j) = 1$) from those that do not (%$\theta_{eq}(r_i)\neq\theta_{eq}(r_j)$, 
$\classifier_{eq}(r_i, r_j) = 0$).  Given an (unlabeled) test set of record pairs $\candidateset_{test}$, the matcher $\classifier_{eq}$ classifies  pairs, aiming for a resolution $\resolution\models\theta_{eq}$. In this setting, a Cross Entropy (CE) loss is used to train (fine-tune) the classifiers~\cite{li2020deep}. Let $y_{ij}$ and $\hat{y}_{ij}$ be the label %(with respect to the equivalence intent) 
and the likelihood score assigned by a matcher to $(r_i, r_j)$, respectively. The cross entropy loss for the record pair $(r_i,r_j)$ is given by: 
%{\small
%\begin{equation}\label{eq:ce}
%\begin{aligned}
%CE((r_i,r_j), \hat{y}_{ij}, y_{ij}) = & \\
% -(y_{ij}\cdot\log(\hat{y}_{ij}) & + (1-y_{ij})\cdot\log(1-\hat{y}_{ij}))
%\end{aligned}
%\end{equation}
%}
\begin{equation}\label{eq:ce}
\begin{aligned}
CE(\hat{y}_{ij}, y_{ij}) = -(y_{ij}\cdot\log(\hat{y}_{ij})  + (1-y_{ij})\log(1-\hat{y}_{ij}))
\end{aligned}
\end{equation}
   
%\end{sloppypar}

Similar to the way matchers learn to resolve a dataset under an equivalence intent, we can utilize the flexibility of learning-based matchers to learn other intents as well. Classifiers learn what we teach them. Thus, by using the same $\candidateset_{train}$ with different labels, we can train a matcher to resolve record pairs for any given intent. Therefore, %Similar to the equivalence intent, we assume that 
given an intent $\pi_{p}$, we can create a respective dataset $\mathcal{D}_{train}(\pi_{p}) = \{(r_i, r_j),y^{p}_{ij}\}_{(r_i,r_j)\in \candidateset_{train}}$, such that $y^{p}_{ij}=1$ if the record pair $(r_i,r_j)$ satisfies $\theta_{p}(r_i)=\theta_{p}(r_j)$, and $y^{p}_{ij}=0$ otherwise.

%We now provide two baseline solutions that address \problemName.

%\subsection{In-parallel: Multiple Matchers For Multiple Intents}\label{sec:multi}
\subsection{Multiple Matchers For Multiple Intents}\label{sec:multi}

\begin{figure}[htpb]
	\centering
	\includegraphics[width=0.75\columnwidth]{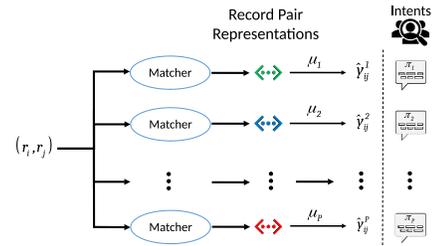}
%	\vskip -0.12in
	\caption{In-parallel: a Binary Matcher per Intent} 
	\label{fig:multi}
\end{figure}

An \emph{in-parallel} approach to \problemNameSpace solves each intent separately, as illustrated in Figure~\ref{fig:multi}. Specifically, an in-parallel approach transforms the multi-label problem into a set of binary problems, one for each label, as suggested by Read {\em et al.}~\cite{read2011classifier}. This means that we solve each of the intents using separate training, yielding a different matcher and different record pair representations for each intent.%, see Figure~\ref{fig:multi}.

%\vskip -0.12in

Given a set of intents $\Pi = \{\pi_{1}, \pi_{2}, \cdots, \pi_{P}\}$ (Section~\ref{sec:MI}), we train $P$ binary matchers (one for each intent), where the matcher $\classifier_{p}$ is responsible for creating a resolution $\resolution_{p}$ to satisfy $\theta_{p}$. We fine-tune each matcher independently using the CE loss (Eq.~\ref{eq:ce}), \emph{i.e.,} $CE(\hat{y}^{p}_{ij}, y^{p}_{ij})$, where $\hat{y}^{p}_{ij}$ is the likelihood score assigned by a matcher to $(r_i, r_j)$ according to the the $p$'th intent and $y^{p}_{ij}$ is the label. 

%In this setting, a Cross Entropy (CE) loss is used to train (fine-tune) each of these classifiers (independently). CE((r_i,r_j), \hat{y}^{p}_{ij}, y^{p}_{ij})

%Given a record pair $(r_i, r_j)$, let $y^{p}_{ij}$ be a the label with respect to the $p$'th intent. In addition, let $\hat{y}^{p}_{ij}$ be the likelihood score assigned by a matcher to $(r_i, r_j)$. The cross entropy loss for $(r_i,r_j)$ is given by: 
%\begin{equation}\label{eq:ce}
%CE((r_i,r_j), \hat{y}^{p}_{ij}, y^{p}_{ij}) = -(y^{p}_{ij}\cdot\log(\hat{y}^{p}_{ij}) + (1-y^{p}_{ij})\cdot\log(1-\hat{y}^{p}_{ij}))
%\end{equation}   
%\end{sloppypar}
%\rs{I need to revise this. We used $\mathcal{\resolution}$ to denote a set of resolutions and not a set of resolution creation mappings.. }
%\ag{please read again carefully to make sure we use the right notation everywhere. I was a bit confused between classifier and resolution creating mapping, which I believe are the same} 
% Given an (unseen) record pair, $\classifier$ is used to provide $P$ labels, one for each intent. 
Given a (test) set of candidate pairs $\candidateset_{test}$, a multi-label matcher $\classifier$ returns $P$ binary labels as a solution for \problemName. A set of resolutions $\mathcal{\resolution} = \{\resolution_{1}, \resolution_{2}, \cdots, \resolution_{P}\}$ is created with $P$ binary matchers $\Pi = \{\pi_{1}, \pi_{2}, \cdots, \pi_{P}\}$, independently applying each $\classifier_{p}\in\classifier$. 

\subsection{Joint Learning of Multiple Intents}\label{sec:prior}
A possible disadvantage of an in-parallel solution is that intents are unaware of each other and the relationships between them are ignored. %Next, we present \modelName, aiming to create an intent-based representation of pairs, but, at the same time, acknowledging that the intents are not independent.
We now lay the groundwork for \modelName, aiming to jointly learn multiple intents. To do so, we treat the multi-label problem directly. The inherent assumption of this approach is that, since intents are interrelated, there exists some (latent) record pair representation (embedding) that can satisfy multiple intents simultaneously. This assumption, in fact, is also beneficial performance-wise. The in-parallel solution (Section~\ref{sec:multi}) requires $P$ (the number of intents) separate training phases, whereas the multi-label solution requires a single, combined, training phase. 

%\begin{sloppypar}
%\end{sloppypar}

\begin{figure}[htpb]
	\centering
	\includegraphics[width=0.75\columnwidth]{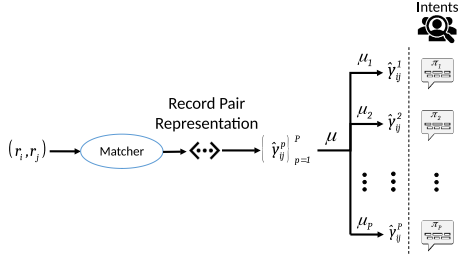}
%	\vskip -0.12in
	\caption{Multi-label Matcher}
	\label{fig:onesize}
\end{figure}

%\ag{shouldn't fig 4 look like the left part of fig 5? perhaps we can have a single figure where the left part is clearly marked as the multi-labeler?}
Given a set of intents $\Pi = \{\pi_{1}, \pi_{2}, \cdots, \pi_{P}\}$, we create a multi-label dataset $\mathcal{D}_{train}(\Pi) = \{(r_i, r_j),(y^{1}_{ij}, y^{2}_{ij}, \dots, y^{P}_{ij})\}_{(r_i,r_j)\in \candidateset_{train}}$, such that $y^{p}_{ij}$ is a binary label corresponding to the intent $\pi_{p}$. Using the multi-label dataset, we train (fine-tune) a single multi-label matcher, $\classifier$, out of which we can also create a binary matcher for each intent, as illustrated in Figure~\ref{fig:onesize}.

Having a multi-label output, the standard cross entropy loss (Equation~\ref{eq:ce}) does not capture the severity of an error per each head. For this reason, we replace it with a multi-label adaptation~\cite{durand2019learning}. Let $\{y^{1}_{ij}, y^{2}_{ij}, \dots, y^{P}_{ij}\}$ and $\{\hat{y}^{1}_{ij}, \hat{y}^{2}_{ij}, \dots, \hat{y}^{P}_{ij}\}$ be the correct and predicted (likelihood scores) multi-label for $(r_i, r_j)$, respectively. The loss for the record pair $(r_i,r_j)$ is given by:
%\begin{scriptsize}
\begin{equation}\label{eq:bce}
\begin{aligned}
BCE((r_i,r_j), &\{\hat{y}^{p}_{ij}\}_{p=1}^{P}, \{y^{p}_{ij}\}_{p=1}^{P}) = \\
\frac{1}{P}\sum_{p=1}^{p=P} & -w_{p}\cdot (y^{p}_{ij}\cdot \log\sigma(\hat{y}^{p}_{ij}) + (1 - y^{p}_{ij})\cdot \log(1-\sigma(\hat{y}^{p}_{ij})))\\
\end{aligned}
\end{equation}
%\end{scriptsize}
where $\sigma(\hat{y}_{ij}) = \frac{1}{1+e^{-\hat{y}_{ij}}}$ is the $sigmoid$ function and $w_{p}$ is a weight, assigned to an error of the $p$'th intent. %It is worth noting that in the experiments we have conducted, $w{p}$ was selected to be 1 for all intents. \ag{why didn't we vary $w{p}$?} %\bg{Is there a reason for not citing the paper of Yarin Gal here?}

%\begin{figure*}[htpb]
%	\centering
%	\includegraphics[width=\textwidth]{./figs/FlexER_small}
%	\caption{\modelNameSpace GCN Inference Illustrated for a Single Record Pair} 
%	\label{fig:flexer}
%\end{figure*}

To offer an intuition to the design choices of Eq.~\ref{eq:bce} we note that a multi-label problem is a generalization of multi class, where only a single class can be correct, \emph{i.e.,} $y^{1}_{ij} + y^{2}_{ij} + \dots + y^{P}_{ij} = 1$. With multi-class classification, an extended version of the CE loss (Eq.~\ref{eq:ce}) typically applies a $Softmax$ transformation over the predictions %to create a distribution over the labels that 
representing a distribution over the classes to be the single correct class. With multiple intents, multiple classes can be labeled as true simultaneously. Therefore, instead of using $Softmax$, we apply a $sigmoid$ activation by element. To compensate for class (intent) imbalance, where some intents create bigger resolution sets (see Table~\ref{tab:Datasetsapp} for imbalance illustration over the datasets in our experiments), intents can be assigned with different weights ($w_{p}$).

%\begin{sloppypar}
Given a (test) candidate record pair from $\candidateset_{test}$, we apply the trained resolution creation mappings over the intents to create a set of solutions $\{\classifier_{1}(r_i, r_j), \classifier_{2}(r_i, r_j), \cdots, \classifier_{P}(r_i, r_j)\}$ out of which we can create a set of resolutions $\mathcal{\resolution} = \{\resolution_{1}, \resolution_{2}, \cdots, \resolution_{P}\}$. 
%\end{sloppypar}

\section{\modelName: Enhanced Resolution with Multiple Intents}\label{sec:flexer}
%The main difference between the solutions presented in sections~\ref{sec:multi} and~\ref{sec:prior} is in the representation of record pairs. In the 
Our proposed flexible entity resolution (\modelName) solution is a flexible approach to the \problemNameSpace problem (Problem~\ref{def:PD}). \modelNameSpace zeros in on the matching phase, casting the problem as a multi-class multi-label task. 
\modelNameSpace builds upon the initial representations drawn from the in-parallel approach (Section~\ref{sec:multi}) and extends the
%Recall that the in-parallel solution (Section~\ref{sec:multi}) learn separately a fine-tuned \emph{intent-based representation} for each intent, returning an independent solution $\hat{y}^{p}_{ij}$ for the $p$-th intent (see Figure~\ref{fig:multi}). The 
multi-label solution (see Section~\ref{sec:prior} and Figure~\ref{fig:onesize}), %creates a single global record pair representation, jointly fine-tuned, and returns a joint solution $\{\hat{y}^{p}_{ij}\}_{p=1}^{P}$, out of which we extract $\hat{y}^{p}_{ij}$ as the solution for the $p$-th intent. 
offering support to learning interrelationships among intents.

Recall that we target intents that are neither given explicitly nor known a-priori. They are not available as categories in the dataset and are not guided by human experts. Rather, they may be the outcome of occasional labeling following either explicit or implicit input of users. In particular, such labeling may follow different parameter tuning of algorithmic solutions to entity resolution that yields multiple interpretations of the data at hand. In such a setting, we argue that solving \problemNameSpace requires a learning component to understand the interrelationships among intents as part of the main general task of yielding intent-level resolutions.

\begin{figure}[t]
	\centering
	\includegraphics[width=0.9\columnwidth]{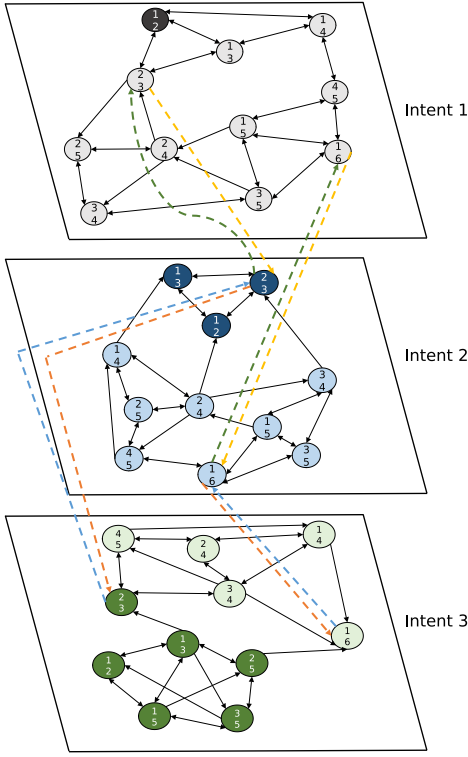}
	\caption{\modelNameSpace graph. Record pairs are represented with numbered circles, such that the same number refers to the same pair for all intents.} 
	\label{fig:flexer}
\end{figure}

To obtain the maximum utility from record pair representations, \modelNameSpace makes use of an \emph{intents graph}, a multiplex graph~\cite{hamilton2020graph} over record pair intent-based representations that codifies intents interrelationships (see Section~\ref{sec:intentrelations}). A {\em multiplex graph} is a special type of a {\em multi-relational graph}~\cite{hamilton2020graph}, a graph with multiple edge types. Formally, a multi-relational graph is a triplet $G=\left( V,E,R\right)$, where $V$ is a set of nodes, $R$ is a set of relation types and $E$ is a set of typed edges that connect pairs of nodes. An edge $\left( v_i, r, v_j\right)$ is a triplet, where $v_i,v_j \in V$ and $r \in R$. %\bg{Roee, Iv'e added an example but i'm not sure that it is helpful here. Our particular case is very different than ordinary cases of multi-relational graphs. Consider if we want to keep this example or not} Relation types vary according to the data the graph was constructed by. For example, in a paper citations dataset the relations can be authored (between nodes representing authors and papers) and cited (between nodes representing papers). 
Multiplex graphs are built in layers (see Figure~\ref{fig:flexer}, to be discussed in Section~\ref{sec:Gcreate}, for iilustration) and every node is duplicated in each of the layers. Each layer %\bg{Roee, this is a good comment. Iv'e added a footnote comment in Section~\ref{sec:Gmessage}. Please take a look} 
represents a unique concept and {\em intra-layer edges} correspond to relationships between nodes according to this concept. Nodes across different layers can be connected using {\em inter-layer edges}. 

Multiplex graph allows an intuitive representation of a problem, of which the information regarding an object (node) is multi-faceted. Such graphs are prevalent in multi-dimensions systems like transportation networks and social networks. We design our \emph{intent graph} as a multiplex graph, where a layer corresponds to an intent, and node in a layer corresponds to a record pair representation, according to that intent.
%\begin{figure}[h]
%	\centering
%	\includegraphics[width=\columnwidth]{./figs/multi_task_multi_label.pdf}
%	\caption{Multi-task Extension of the Multi-label Solution (Figure~\ref{fig:onesize}): producing both \textbf{global} and \textbf{intent-based} record pair representations.} 
%	\label{fig:multi_task}
%\end{figure}
%\modelName, proposed herein, enhances the multi-label solution. For illustration purposes, consider Figure~\ref{fig:multi_task}, where a matcher is trained to create a global record pair representation with a side effect of a joint solution $\{\hat{y}^{p}_{ij}\}_{p=1}^{P}$. The global representation is then utilized to create individual intent-based representations and a respective solution $\hat{y}^{p}_{ij}$. 
% we apply a powerful graph inference mechanism to construct an \emph{intent graph} 
The intent graph consists of $P$ node representation layers. We assign each record pair $(r_i,r_j)\subseteq C$ with a corresponding set of nodes $\left\lbrace v^1_{ij}, v^2_{ij}, \cdots, v^P_{ij}\right\rbrace$, built upon initial representations drawn from the in-parallel approach (Section~\ref{sec:multi}). Then, nodes referring to the same record pair are connected via intra-layer edges, whereas nodes within the same layer are connected through inter-layer edges to their closest record pair counterparts. 

By applying a graph neural network (GNN) model over it, \modelNameSpace provides enriched intents-aware representations, which results in improved resolutions, yielding an intent-aware prediction for all record pairs.
The entire process of \modelNameSpace involves three main phases, namely {\em graph creation, message propagation}, and {\em prediction per intent}, as detailed next.

%The \emph{intent graph} is fed into a graph Graph neu (GCN). Therefore, given an input record pair $(r_i,r_j)$, an undirected graph $G_{ij}=\left( V_{ij},E_{ij}\right)$ is created. The nodes of $G_{ij}$ ($V_{ij}$) correspond to the intent-based representations and the edges of $G_{ij}$ ($E_{ij}$) to the interrelations among the intents. 
%\modelNameSpace GCN inference follows~\cite{kipf2016semi} and involves three main phases, namely the graph creation, messages propagation, and prediction per intent, as detailed next. 
%\subsection{Preliminaries}% -- Record Pair Representations and Intent Inter-relationships}
%\label{sec:preliminaries}
%Before describing \modelName, we provide the necessary background to better understand its building blocks.

%\bg{Do you think that short preliminaries about GNN is required? And what about vector-space similarity?}

\subsection{Graph Creation}% -- Record Pair Representations and Intent Inter-relationships}
\label{sec:Gcreate}
We begin with a brief description of generating intent-based representations, which are used to initialize the nodes of the graph, after which we discuss edge creation and labeling.

%\begin{figure*}[htpb]
%	\centering
%	\includegraphics[width=\textwidth]{./figs/Multiplex.eps}
%	\caption{\modelNameSpace graph. Record pairs are represented with numbered circles, such that the same number refers to the same pair for all intents.} 
%	\label{fig:flexer}
%\end{figure*}
%\begin{figure*}[t]
%	\centering
%	\includegraphics[width=1.0\textwidth]{./figs/Multiplex_4_layers.eps}
%	\caption{\modelNameSpace graph. Record pairs are represented with numbered circles, such that the same number refers to the same pair for all intents.} 
%	\label{fig:flexer}
%\end{figure*}

%\begin{figure}[h]
%	\centering
%	\includegraphics[width=0.8\columnwidth]{./figs/matcher_per_intent.pdf}
%	\caption{In-parallel: a Binary Matcher per Intent} 
%	\label{fig:multi}
%\end{figure}

We use a pre-trained model and fine-tune it %A unique characteristic of the deep learning-based matchers is that the record pair representation (see Section~\ref{sec:ER}) can be fine-tuned according to a specific task. This becomes especially beneficial in our setting as we aim to create a record pair representation tuned 
for each intent (intent-based representation). To support intent interrelationships, %While intent-based representations can be derived for independent learning of single intents (Section~\ref{sec:multi}), we note that intents are interlaced and hence, 
we enrich intent-based representations with the aid of other intent-based representations. 
	
\subsubsection{Nodes: }
\label{sec:Gnodes}
First, \modelNameSpace constructs a set of intent-layer nodes $V=\left\lbrace V^1, V^2, \cdots, V^P\right\rbrace$ (see Figure~\ref{fig:flexer} for an illustration of the constructed graph). These intent layers are initialized with the independent intent-based representation of all record pairs in a given set of candidate pairs $C$, namely $V^p = \left\lbrace v^p_{ij}, \forall (r_i,r_j) \in C\right\rbrace$. In total, the number of nodes in the intents graph is $\left| C\right| \cdot \left| P\right|$.

We treat a node representation of a record pair as an initial feature vector, capturing the underlying semantics of the pair with respect to the task for which it was trained or tested against. Specifically, this task involves predicting whether a record pair matches a given intent. Node representations of the same pair for different intents are obtained independently, as the training process is carried out separately for each intent (see Section~\ref{sec:multi}).
As a result, node representations of different intent layers, while having the same dimensionality, belong to different latent spaces and their node representations %that where trained on different tasks (in our case, different intents) 
are not aligned with each other, as the $i^{th}$ feature of each intent carries different meanings for the same record pair. Therefore, a main challenge here involves offering a meaningful way for pairs to update their initial representation by communicating with nodes at different layers. 

%Node representations can benefit from their same-pair yet different intent peers.
\begin{sloppypar}
\subsubsection{Inter-Layer Edges: }
\label{sec:Ginteredges}
To pave the way to intents-aware node representation, we define a set of inter-layer edges $E_{inter}=\left\lbrace E^{p, \tilde{p}}, \forall \left( p, \tilde{p}\right)  \in P \times P \right\rbrace$ where $p \neq \tilde{p}$. A set $ E^{p, \tilde{p}} \in E_{inter}$ contains exactly $\left| C\right|$ edges, connecting a node from intent $p$ to its {\em peer}, the node representing the same record pair, of intent $\tilde{p}$. These edges are utilized to propagate representation changes among layers. It is noteworthy that $E^{p, \tilde{p}}$ and $E^{\tilde{p}, p}$ are different and directionality of message propagation plays a key role in the next phase (Section~\ref{sec:Gmessage}).
The total number of inter-layer edges is $\left| C\right| \cdot \left| P\right| \cdot \left| P-1\right|$.
%\ag{I started rewriting but then realize I'm not sure this is at all needed since we have a clear semantics to what edges mean. Consider eliminating the following:} Recall that the model is unaware of intent meanings or their semantics.  connection between intents (\emph{e.g.,} subsumption); therefore, we connect by edges every pair of intent layers, assuming that meaningful inter-intents relationships will be learned by the model.
\end{sloppypar}
\subsubsection{Intra-Layer Edges: }
\label{sec:Gintraedges}
In addition to inter-layer edges, we also define $E_{intra}=\left\lbrace E^{p}, \forall p \in P \right\rbrace$, a set of intra-layer edges connecting a given node to its closest counterparts among the nodes of the same intent. This way, a pair representation can be enriched with those that are closely related to it for a specific intent. \update{The intuition behind adding intra-layer edges comes from the $k$ nearest neighbor algorithm, where vector-space similarity is interpreted as agreement between samples~\cite{bhatia2010survey}. In Section~\ref{sec:intra_layer_analysis} We empirically show that adding intra-layer edges improves \modelName's performance.}
%\bg{Before the next sentence, refer to the relevant part of the introduction} Node representations of the same intent tend to concentrate in the same areas of the vector-space; Moreover, within such a cluster, nodes tend to form a more condensed bundle (\bg{see Figure ... (refer to the relevant part)}) with other nodes with which they agree, regarding the task at hand. In accordance with this observation, 
%Nodes update their representation by receiving messages from their own intent-layer counterparts.
 
We connect a node to its $k$ nearest neighbors, computed over an initial node representation. \update{The identity of a node's nearest neighbors may change during the iterative GNN process, yet the set of edges is predetermined and remains constant, which is in line with common good practices in GNN construction.} The total number of generated intra-edges is $\left| C\right| \cdot \left| P\right| \cdot \left| k\right|$. Note that intra-edges are directional. Therefore, while $v^p_{ij}$ might be among the $k$ nearest neighbors of $v^p_{st}$, the opposite does not necessarily hold. 

\subsubsection{Intent Graph as a Whole: }
We illustrate the graph structure using Figure~\ref{fig:flexer}. In this example there are 11 record pairs, taken from Table~\ref{tab:example}. A pair in this figure is denoted by a node with the numbers of the two records it represents, and it appears in each of the $3$ given intent-layers. As portrayed by this figure, nodes can be spread differently in the space for different intents. Within a layer, a node is connected to its $k$ ($k=3$ in this case) nearest neighbors with incoming edges. In addition, a node from one layer is connected to its peers from all other layers. For sake of ease of presentation, we present the inter-layer edges for the record pairs $\left( 2,3\right)$ and $\left( 1,6\right)$ only, and only between consecutive layers. We use a bidirectional edge to signify that two nodes are among the $k$ nearest neighbors of each other, {\em e.g.}, pairs $(r_1,r_5)$ and $(r_1,r_6)$ for the \emph{Intent $1$} layer. When a record pair is a match according to a given intent, its underlying intent-layer node is marked with a dark color and white numbering, {\em e.g.}, the pair $(r_2,r_3)$ for the \emph{Intent $3$} layer.

\subsection{Message Propagation}
\label{sec:Gmessage}
%Our suggested graph structure is illustrated in Figure~\ref{fig:flexer}. To convey the idea of linking intents, layers are graphically separated, and the order of pair record nodes remains constant over all intent. 
The multiplex graph, whose generation is detailed in Section~\ref{sec:Gcreate}, lays the foundation to the usage of a GNN model as a mechanism to learn to characterize intents, as elaborated next.
In what follows, it is worth noting that the term layer, as is being used in GNNs, refers to the pipeline of receiving messages from neighboring nodes, aggregating them, and applying a fully connected neural network with an activation function. For the intent graph, being a multiplex graph, a layer refers to a set of nodes of the same intent.

A general GNN architecture is composed of $q$ layers, each producing a hidden state vector, is generated by aggregating the vectors of adjacent nodes. Using the multi-layer GNN, each node iteratively transmits its current information to itself and its neighboring nodes (connected by outgoing edges). Numerous GNN models has been introduced in recent years. In this work, we follow the model of GraphSAGE~\cite{hamilton2017inductive} due to its popularity and ease of use.

The first hidden layer for the node $v$, $h_{v}^{(0)}$, is initialized to be the intent-based representation of the respective intent. %(see Section~\ref{sec:Gcreate}). 
In GraphSAGE, the update to the hidden state vectors at layer $t+1$ ($t\in\{0,1,\dots,q-1\}$) of node $v$ is executed in two stages. First, neighbor messages are aggregated, yielding $v$'s neighborhood representation as follows.
\begin{equation}\label{eq:Naggregation}
h^{(t+1)}_{N(v)} = AGG_{t+1}\left( \left\lbrace h^t_u, \forall u \in N(v) \right\rbrace \right)  
\end{equation}
where $AGG_{t+1}$ is an aggregation function (\emph{e.g., } sum or mean) and $N(v)$ is the set of $v$'s neighbors, connected by incoming edges. Then, this representation is concatenated to $v$'s previous layer representation, such that the resulted vector is fed into a fully connected neural network, followed by an activation function, as follows.
%traditionally given by:
\begin{equation}\label{eq:GraphSAGELayer}
h^{(t+1)}_{v} = \sigma\left( W^t \cdot CONC  \left( h^{(t)}_{v}, h^{(t+1)}_{N(v)}\right) \right)  
\end{equation}
where $CONC$ is a vector concatenation operator, $\sigma$ is an activation function (\emph{e.g.,} \emph{ReLU}) applied in each layer except the last, and $W^{(t)}$ is the weights matrix of the $t$-th convolution layer.
%However, we deal with a
%A multiplex graph is a special case of a heterogeneous graph of which different types of nodes and edges exist. Therefore, we 
We use a modified version of Eq.~\ref{eq:Naggregation}, adjusted for multiplex graphs. For more details we refer the interested reader to~\cite{schlichtkrull2018modeling}.

%\begin{equation}\label{eq:edge}
%h_{v}^{(t+1)} = \sigma\left(\sum_{v^{\prime} \in N(v) \cup v}\frac{1}{c_{v,v^{\prime}}}\cdot h_{v}^{(t)}\cdot W^{(t)}\right) 
%\end{equation}
%where $\sigma$ is an activation function (\emph{e.g.,} \emph{ReLU}) applied in each layer except the last, $W^{(t)}$ is the weights matrix of the $t$-th convolution layer, and $N(v)$ is the set of neighbors of the node $v$ in $G$. $c_{v,v^{\prime}}$ is a normalization factor of the edge from source node $v^{\prime}$ to target node $v$ ($e_{v,v^{\prime}}$), incorporating the corresponding feature vector.\footnote{see~\cite{kipf2016semi} for additional details}

\subsection{Prediction per Intent}
\label{sec:Gpredict}

The final phase of training, after messages are propagated through the GNN, involves obtaining a prediction for each intent over all final hidden layers. \modelNameSpace is trained over $P$ versions of the same graph, one for each intent, to allow proper fine-tuning with respect to the target intent. %In what follows, we now describe how to 

To provide a prediction for an intent $\pi_{p}$,
the final hidden representation of the node $v^{p}_{ij}$ ($h_{v^{p}_{ij}}^{(q)}$), corresponding to the intent $\pi_{p}$, is fed into a fully connected layer followed by a $softmax$ and an $argmax$ operations yielding the following:
\begin{equation}
\classifier_{p}(r_i,r_j)= {\mathrm{argmax}}\left(softmax \left( W^{(fn)}\cdot h_{v^{p}_{ij}}^{(q)}\right)\right)
\end{equation} 
%\begin{sloppypar}
%	\noindent 
where $W^{(fn)}$ is the weights matrix of the fully connected layer and $softmax( W^{(fn)}\cdot h_{v^{p}_{ij}}^{(q)})$ is a 2-dimensional prediction vector with entries corresponding to classes likelihood (its second entry, corresponding to the label $1$, can be used as a likelihood score for $(r_i,r_j)$, see Section~\ref{sec:ER}). Given a record pair $(r_i,r_j)$, $\classifier_{p}(r_i,r_j)$ is assigned with the entry ($0$ or $1$) that has the highest likelihood, serving as the prediction for the intent $\pi_{p}$. 
%For example, in Figure~\ref{fig:flexer} the representation of $(r_i,r_j)$ with respect to $\pi_{1}$ (result of top intent, in green) is used in the intent graph that is modified using the GCN to provide a prediction $\classifier_{1}(r_i,r_j)$.

%\input{GranularER}
\section{Empirical Evaluation}\label{sec:eval}

%\rs{revisit:} We conducted a set of experiments to test \modelName's ability to offer accurate resolutions and improve on single intent resolutions. We begin by describing the benchmarks (Section~\ref{sec:benchmarks}) and detailing the experimental setup (Section~\ref{sec:setup}). The main results, detailed in Section~\ref{sec:res}, can be summarized as follows.
\update{We conducted a set of experiments to test \modelName's ability to offer accurate resolutions and improve on single intent resolutions. We also evaluate the benefit of learning intents over intent graphs using a GNN architecture.} We begin by describing the benchmarks (Section~\ref{sec:benchmarks}) and experimental setup (Section~\ref{sec:setup}). Our main results, can be summarized as follows.

%\begin{sloppypar}
%	\begin{compactitem}
	\begin{itemize}
		\item \emph{\modelNameSpace effectively solves the task of \problemNameSpace} (Section~\ref{sec:res_multi}). 
		\item \emph{\modelNameSpace outperforms state-of-the-art baselines} over all examined benchmarks for the \er task  (Section~\ref{sec:res_universal}).
		\item \update{\emph{\modelNameSpace benefits from inter-layer edges}, making use of intent-based information to improve its performance for the task of universal \ernospace~(Section~\ref{sec:inter_layer_analysis}).}
		\item \update{\emph{\modelNameSpace utilizes the component of intra-layer edges}, such that connecting all nodes with other nodes from their own multiplex graph layer enhances the performance of the model over the task of universal \ernospace~(Section~\ref{sec:intra_layer_analysis}).}
%	\end{compactitem} 
	\end{itemize}
%\end{sloppypar}

%\noindent In addition, \update{Section~\ref{sec:inter_layer_analysis} analyzes the benefit of intent interrelationships in solving universal \ernospace\. In particular, Section~\ref{sec:Interrelationships} shows the benefit of learning intents over intent graphs using a GNN architecture. Section~\ref{sec:intra_layer_analysis} analyze the design choice of number of intra-layer edges per node.}
\subsection{Benchmarks and Intent Definition}\label{sec:benchmarks}

We experimented with three benchmarks, among which a new publicly available benchmark for the \problemNameSpace task (AmazonMI). In addition, we conducted experiments with a standard (clean-clean) entity matching benchmark (Walmart-Amazon)~\cite{konda2016magellan} and a product matching benchmark (WDC)~\cite{primpeli2019wdc}. The datasets and their respective intent-based labels are provided in a git repository~\footnote{\url{https://github.com/BarGenossar/FlexER/tree/main/data/}}.%\footnote{\url{https://github.com/authorAnonymousGit/FlexER/}}

\begin{table}[htpb]
	\caption{Benchmark datasets used in the evaluation}
	\label{tab:Datasets}
	\begin{center}
		\scalebox{0.9}{\begin{tabular}{lccc}
				\hline
				\textbf{Dataset} & \textbf{\#Records} & \textbf{\#Pairs} & \textbf{\#Intents} \\
				& ($|D|$) &   ($|\resolution|$) &  ($|\Pi|$) \\\hline
				AmazonMI & $3,835$ & $15,404$ & $5$ \\
				Walmart-Amazon & $24,628$ & $10,242$ & $4$  \\
				WDC & $10,935$ & $30,673$ & $2$ \\\hline
		\end{tabular}}
	\end{center}
\end{table}

%\begin{table}[t]
%	\caption{Benchmark datasets used in the evaluation}
%	\label{tab:Datasets}
%	\begin{center}
%		\scalebox{1.25}{\begin{tabular}{lccc}
%				\hline
%				\textbf{Dataset} & \textbf{\#Records} & \textbf{\#Pairs} & \textbf{\#Intents} \\
%				& ($|D|$) &   ($|\resolution|$) &  ($|\Pi|$) \\\hline
%				AmazonMI & $3,835$ & $15,404$ & $5$ \\
%				iTunes-Amazon & $62,830$ & $539$ & $4$  \\
%				Walmart-Amazon & $24,628$ & $10,242$ & $4$  \\
%				WDC & $10,935$ & $30,673$ & $2$ \\\hline
%		\end{tabular}}
%	\end{center}
%\end{table}	

Metadata including cardinality of $D$ (\#Records), $\resolution$ (\#Pairs), and number of intents (\#Intents) is provided in Table~\ref{tab:Datasets}.
Table~\ref{tab:Datasetsapp} presents the proportion of positive (matching) samples (\%Pos) for each intent over the training, validation, and test sets. %Following previous works~\cite{konda2016magellan,mudgal2018deep,li2020deep}, each dataset was split into train/validation/test sets with a 3:1:1 ratio.

\begin{table}[htpb]
	\caption{Positive labels ($1$) proportion by dataset and intent}
	\label{tab:Datasetsapp}
	\begin{center}
		\scalebox{0.9}{\begin{tabular}{llccc}
				\hline
				\textbf{Dataset} & \textbf{Intent} & \multicolumn{3}{c}{\textbf{\%Pos}} \\
				& & \textbf{Train}  & \textbf{Valid} & \textbf{Test}\\\hline
				\multirow{5}{*}{\begin{tabular}[c]{@{}c@{}}AmazonMI\end{tabular}} &  (1) Eq. & 15.1\% & 16.2\% & 15.4\%\\
				&  (2) Brand & 20.0\% & 21.3\% & 21.4\%\\
				&  (3) Set-Cat & 49.7\% & 50.7\% & 49.0\%\\
				&  (4) Main-Cat. & 66.8\% & 67.3\% & 67.2\%\\
				&  (5) Main-Cat. \& & & &\\
				&   \hspace{3.55mm} Set-Cat. & 49.7\% & 50.7\% & 49.0\%\\\hline
				\multirow{4}{*}{\begin{tabular}[c]{@{}c@{}}Walmart-Amazon\end{tabular}}  & (1) Eq. & 9.4\% & 9.4\% & 9.4\%\\
				& (2) Brand & 75.7\% & 75.7\% & 76.4\% \\
				& (3) Main-Cat. & 79.9\% & 79.0\% & 80.0\% \\
				& (4) General-Cat. & 89.7\% & 90.2\% & 90.5\% \\
				\hline
				\multirow{2}{*}{\begin{tabular}[c]{@{}c@{}}WDC\end{tabular}} & (1) Eq. & 11.6\% & 11.4\% & 11.3\%\\
				& (2) Cat. & 43.8\% & 43.8\% & 43.8\%\\
				& \update{(3) General-Cat.} & \update{67.0\%} & \update{66.6\%} & \update{67.2\%} \\\hline
		\end{tabular}}
	\end{center}
\end{table}

The datasets and corresponding intents %for each dataset 
are detailed next. %illustrated in Figure~\ref{fig:intents}. 
All intents overlap (see Definition~\ref{def:overlap}) and some are subsumed (see Definition~\ref{def:subsum}). %are connected with an arrow from the supersuming intent to the subsumed intent. 
It is worth noting that \update{Walmart-Amazon and WDC} benchmarks are labeled solely for equivalence. Therefore, to support the empirical evaluation, we have created additional labeling using the provided data and known (to us but not the model) relationships between intents. Specifically, we ensure that labels according to a supersuming intent comply with subsumed intents. 
%For example, if a pair is labeled as $1$ for the {\em main category} intent (Walmart-Amazon dataset), it should also be labeled as $1$ for the {\em General Category} intent. Additional details regarding the datasets and the labeling are given next. 

%Using the subsumption of intents (see Definition~\ref{def:subsum}), over the standard datasets (iTunes-Amazon and Walmart-Amazon) labels of supersumed were modified to comply with the subsumed intent. For example, if a pair is labeled as $1$ for an ``same album'' intent (iTunes-Amazon dataset), it will be automatically labeled as $1$ for the ``same artist'' intent. All labels are modified to comply with the equivalence intent. Additional details regarding the dataset are given as follow.

%\vspace{.1cm}
%\noindent\textbf{AmazonMI:} 
%\vspace{.1cm}
%\noindent\textbf{The \emph{amazon multi intent} benchmark (AmazonMI)}
\par{\textbf{The \emph{amazon multi intent} benchmark (AmazonMI)}}
 consists of $3,835$ products, extracted from the amazon website. Product details include \textsf{asin} (unique identifier), title, brand, and an ordered category set. We only use product titles for matching, while other attributes (including the categories) are used for labeling only. To obtain the set of candidate pairs $\resolution$ we use a standard blocker~\cite{konda2016magellan}\footnote{\url{https://github.com/anhaidgroup/py_entitymatching}}, preserving record pairs that share at least a 4-gram. We have created a ground truth of five intents for this benchmark, namely, \emph{equivalence (Eq.), same brand (Brand), same main-category (Main-Cat.), similar category-set (Set-Cat.), same main-category and similar category-set (Main-Cat. \& Set-Cat.)}, as detailed next. 

Tuple pairs were labeled according to the equivalence intent using an available online list of duplicated products~\footnote{\url{http://deepyeti.ucsd.edu/jianmo/amazon/metaFiles/duplicates.txt}}. %\footnote{\url{http://deepyeti.ucsd.edu/jianmo/amazon/metaFiles/duplicates.txt}} 
For other intents we use the available product metadata. Pairs complying with the \emph{same brand} intent (Brand) share a complete correspondence under the brand attribute. It is worth noting that during preprocessing, we defined the category of books as \emph{book} and electronic books (Kindle) as \emph{Kindle}, since those products did not have a brand value. The main-category ({\em Main-Cat.}) is defined as the first category in the (ordered) category-set provided by Amazon. %The category-set is hierarchical, reflecting the product path in Amazon website. 
The category-set reflects the product path in Amazon's website; Hence, the first (main) category in the category-set depicts the most general affiliation of the product, wheres the last category is the most fine-grained. We define the intent of \emph{similar category set} ({\em Set-Cat.}) as achieving a jaccard similarity of at least $0.4$ between the category-sets of a record in a pair. The last intent ({\em Main-Cat. \& Set-Cat.}) satisfies both same main category and similar category set. %It is important to note that the category-set is not available for the algorithm.

%\vspace{.1cm}
%\noindent\textbf{iTunes-Amazon\footnote{\url{pages.cs.wisc.edu/~anhai/data1/deepmatcher_data/Structured/iTunes-Amazon/}} and Walmart-Amazon\footnote{\url{pages.cs.wisc.edu/~anhai/data1/deepmatcher_data/Structured/Walmart-Amazon/}}:}
%\vspace{-.1cm}
%\vspace{.1cm}
%\begin{sloppypar}
\par{\textbf{Walmart-Amazon}~\footnote{\url{http://pages.cs.wisc.edu/~anhai/data1/deepmatcher_data/Structured/Walmart-Amazon/}} %\footnote{\url{http://pages.cs.wisc.edu/~anhai/data1/deepmatcher_data/Structured/Walmart-Amazon/}}} 
matches entities of various product domains~\cite{konda2016magellan}. %To allow a fair comparison with former works, we 
We use the published pre-defined candidate pair sets provided in the literature as a basis for a ground truth of four intents. \update{Similar to the AmazonMI dataset, the input of the model includes only the title, using other attributes for labeling intents. For this reason, the results we present for the equivalence (Section~\ref{sec:res_universal}) intent cannot be directly compared with those reported in~\cite{li2020deep}.}

The intents used for Walmart-Amazon are \emph{equivalence (Eq.), same brand (Brand), same main category (Main-Cat.)}, and \emph{same general category  (\em General-Cat.)}. The pairs were labeled as \emph{same brand} if the two corresponding attribute values overlap. 
The creation of the category-based intents \emph{Main-Cat.} and \emph{General-Cat.} labels was not straightforward. Differences of category-tagging conventions between the two sources from which the data was taken (Walmart and Amazon) prevent direct comparison between tuples with respect to the \emph{category} attribute. For example, for the matching tuples \emph{targus red tg-6660tr tripod with 3-way panhead} and \emph{new-targus red tg-6660tr tripod with 3-way panhead 66 - meytg6660tr} (only tuple titles are presented here), the category of the first tuple is \emph{photography - general}, whereas the category of the second is \emph{tripods}. Although these tuples match, the category of the first is more general than the second. To align between different category conventions, we manually created a hierarchical list of categories. %, such that each main category is a subset of a more general category. 
The most general categories are \emph{electronics, personal equipment, house} and {\em cars}. Other categories describe more specific concepts, {\em e.g.}, %under the general category (For example, 
\emph{computers} is a subset of \emph{electronics}. The full lists can be found in the git.\footnote{\url{https://github.com/BarGenossar/FlexER/}}
%\end{sloppypar}

%\vspace{.1cm}
\par{\textbf{The web data commons (WDC) dataset}~\footnote{\url{http://webdatacommons.org/largescaleproductcorpus/v2/index.html}}\label{fn:wdc} %\footnote{\url{http://webdatacommons.org/largescaleproductcorpus/v2/index.html}}
 contains product data extracted from multiple e-shops, split into four categories, namely \update{computers, cameras, watches, and shoes}. Following~\cite{li2020deep}, we only use product titles. For fair comparison, we use the pre-defined candidate pair sets, labeled for equivalence intent~\footnote{\url{https://github.com/megagonlabs/ditto/tree/master/data/wdc/all}} %\footnote{\url{https://github.com/megagonlabs/ditto/tree/master/data/wdc/all}} 
 and use, to comply with other dataset sizes, the small size training set. 

We define a {\em category} intent for this dataset, whose labels are obtained by the association to the relevant sub-datasets, each provided in a separate file.%$^{\ref{fn:wdc}}$ 
This structure creates a scenario where all pairs are labeled as matching (1) for the {\em category} intent within a provided category file. Thus, we expanded the dataset with cross-category pairs. To do so, we performed an additional blocking phase to generate additional non-matching pairs. For each pair of categories, we applied a blocker and pairs with a common 4-grams threshold~\cite{konda2016magellan} serve as cross-category, yet similar, tuple pairs. Out of these pairs, we randomly sampled $\sim1,500$ pairs to be added to the original record pairs, yielding a new dataset with $30,673$ record pairs. \update{We created an additional intent of {\em general category} by merging the computers and cameras categories into an electronics category, while watches and shoes were merged into a dressing category.}

Finally, we randomly re-split the dataset into training, validation, and test subsets with the ratio of 3:1:1. Note that the original dataset contained $13,436$ total tuple pairs split into training, validation, and test set with sizes of $7,230$, $1,830$, and $4,398$, respectively. This change affects our ability to reproduce the results reported in~\cite{li2020deep}. The new version of the dataset in provided in the git repository~\footnote{\url{https://github.com/BarGenossar/FlexER/tree/main/data/WDC/WDC_small}}.%\footnote{\url{https://github.com/authorAnonymousGit/FlexER/tree/main/data/WDC/WDC_small}}

\update{\noindent\textbf{Benchmarks Profiling:} The AmazonMI forms a natural environment to the \problemName~problem, while the other two were modified to fit it. Recent works~\cite{primpeli2020profiling, wang2021machamp} focus on characterizing entity matching baselines, based on which we can derive insights regarding our expectations and room for improvements over these benchmarks (with respect to the equivalence intent). Primpeli {\em et al}.~\cite{primpeli2020profiling} defined a set of five profiling criteria, namely schema complexity (number of relevant attributes), textuality, sparsity (existence of null values), development size, and corner cases. Walmart-Amazon and WDC were already analyzed in this work, the former classified as textual with few corner cases, and the latter classified as textual with many corner cases group. Note that while we use a slightly different version of WDC %, since they considered the large size version of each category separately. However, 
	this difference does not change the analysis. %AmazonMI can be placed in between the other two datasets. %two groups suggested in the paper. We observed that 
	AmazonMI consists of a single attribute (title) with no null values. It contains long strings (>7 words) and low number of corner cases (<0.15). It is therefore classified as 1) dense data with simple schema, and 2) textual data with few corner cases. As pointed out by Primpeli {\em et al}.~\cite{primpeli2020profiling}, these groups are expected to be relatively easy to solve; therefore, achieving a substantial improvement over existing baseline is challenging. We present in section~\ref{sec:measures} an evaluation measure (Eq.~\ref{eq:residual_error}) which is aimed to asses the performance of a new model comparing to a given baseline in such cases.}
	
%	\ag{lost you from here forward. try to explain differently:} To summarize, AmazonMI is expected to achieve relatively high scores with the used baseline, such that a substantial improvement in the results would be considered as even more impressive achievement than standard classification task. The same holds for Walmart-Amazon as its group is also considered as relatively easy for matching.}

%\vspace{.1cm}
%It is worth noting that the metadata used for labeling is not available for the algorithm.
%\rs{this needs to be rewritten:} We created, for each category, two tables, namely \emph{A} and \emph{B}, such that the first was composed with the left hand tuples in the record pairs set, and the second with the right hand ones. Next, given a pair of categories we separately used the blocker of~\cite{konda2016magellan},\footnote{\url{https://github.com/anhaidgroup/py_entitymatching}} with at least a common 4-grams threshold on the corresponding tables (\emph{A} and \emph{B}) of the categories. For the $2 \cdot \binom{4}{2}$ pairs sets we randomly sampled at approximatly $1,500$ pairs and combined it with the original pairs (after merging them together, ignoring the original partition), yielding a new dataset with $30,673$ record pairs. As a final step, we randomly split the data set into training, validation and test with the ratio of 3:1:1.

\subsection{Experimental Setup}\label{sec:setup}
Experiments were performed on a server with 2 Nvidia Quadro RTX 6000 and a CentOS 6.4 operating system. Networks were implemented using PyTorch~\cite{paszke2019pytorch} %\footnote{\url{https://pytorch.org/}} 
and PyTorch Geometric~\cite{fey2019fast}. The DITTO matcher~\cite{li2020deep} (see Example~\ref{ex:ditto}) was used to generate vector-based representations for all benchmarks, using RoBERTa~\cite{liu:2019roberta} following~\cite{li2020deep}. The complete code is available in a git.\footnote{\url{http://github.com/BarGenossar/FlexER/}} %\note{say if we experiment with additional BERTs in Amazon.}
%\note{Following~\cite{li2020deep}, since we only deal with large training sets, DistilBERT in DITTO across all settings for faster training}. 

\subsubsection{Implementation Details}
\label{sec:implementation}

%\rs{Bar, please revise:}
We now provide implementation details for the central components in \modelName. We begin with describing the basic matching method (based on DITTO~\cite{li2020deep}), which is used as a preliminary step to create record pair representations, and continue with the graph creation, followed by intent interrelationships learning (using GNN).

%\subsubsection{DITTO}
%\label{sec:dittoImp}
%\vspace{.1cm}
\noindent\textbf{DITTO:}
We adopted the implementation of DITTO~\cite{li2020deep} provided in publicly available code~\footnote{\url{https://github.com/megagonlabs/ditto}}}. %\footnote{\url{https://github.com/megagonlabs/ditto}} 
Following~\cite{li2020deep}, we set the input sequence to 512 tokens, the learning rate to $3e-5$, the number of epochs per run to $15$, and use a batch size of 16.
We tested three optimizations~\cite{li2020deep}, namely data augmentation with the option of deleting spans of tokens, injecting domain knowledge on products, and data summarization to handle long input sequences, of which only the first yielded improved results. Therefore, we report only on the performance with data augmentation. Finally, we report the average performance over 5 different seeds.
For the multi-label baseline (Section~\ref{sec:prior}), we allocate an equal weight for each intent, after preliminary experiments showed no significant difference between this simple heuristic to parameters learning.

%\vspace{.1cm}
\noindent\textbf{Multiplex Graph:} Tuple pair representations serve in generating a multiplex graph that captures intent interrelationships. %To create intra-layer edges, 
We employed vector-space similarity search using \emph{Faiss}~\cite{johnson2019billion} to connect each node to its $k$ nearest neighbors using $L_2$ distance. We tested \update{and report on possible $k$ values ($k\in\left\lbrace 0,2,4,6,8,10\right\rbrace$). For some of the experiments, we report only on the $k$ value} that achieved the best performance over the validation set. %\bg{Discuss with Avi about the analysis of k}

%\vspace{.1cm}
\noindent\textbf{GNN:}
We enrich intent-based tuple pair representations, for better inference, using GNN that learns the interactions between intents (Section~\ref{sec:flexer}). We experimented with two \update{or three} layers of GraphSAGE~\cite{hamilton2017inductive}, noting that more layers slowed down the training without enhancing performance.
We trained the model over 150 epochs using \emph{Adam} optimizer~\cite{kingma2014adam} with learning rate of $0.01$, weight decay of $5e-4$ and cross entropy loss function. We perform the hidden layer update (see Eq.~\ref{eq:GraphSAGELayer}) with \emph{ReLU} activation function. In our experiments we \update{examined several $h_{1}$ dimensions within the set $\left\lbrace 100,150,200,250,300,350,400,450,500 \right\rbrace$, while for three-layer model the dimensionality is set as half of $h_{1}$}. We select the best preforming model over the validation set, and report the results over the test set. 

%It is noteworthy that, with adequate resources, several DITTO models can be trained simultaneously. \bg{Consider to remove this patr, as I will add running time analysis to Section 5}\ag{agree} In addition, the time complexity of creating the multiplex graph and training the GNN are negligible compared to training DITTO. Thus, the overhead of applying \modelName\ is not significantly higher than DITTO, allowing its usage even for large scales of data.

\subsubsection{Methodology}\label{sec:method}

%The methodology of \modelNameSpace follows Section~\ref{sec:flexer}. 
For each benchmark (Section~\ref{sec:benchmarks}), we first train a DITTO-based matcher/s. To generate independent intent-based representations, we train $P$ separate matchers, extracting for each an intent-based representation using the independent $[cls]$ \update{latent} representations (see Example~\ref{ex:ditto}). For the multiple intent representation, we train a multi-task network, including a multi-label output and $P$ binary outputs (one per intent), using a single DITTO matcher. After fine-tuning the multi-task %(M-T) 
network, we extract the intent-based representations, using the latent representation of the layer prior to the output, per intent. These representations are used to construct the intent graph %(Section~\ref{sec:Gcreate}) 
and the rest of the GNN inference (Section~\ref{sec:flexer}). In sections~\ref{sec:res_multi} and~\ref{sec:res_universal} we report on the results of \modelNameSpace using the independent intent-based representations.

%The methodology of \modelNameSpace follows Section~\ref{sec:flexer}. For each benchmark (Section~\ref{sec:benchmarks}), we first train a multi-task network, including a multi-label output and $P$ binary outputs (one for each intent), using a single DITTO matcher. After fine-tuning the multi-task network, we extract the intent record pair representation for each intent, using the latent representation of the layer prior to the output, per intent. These representations are then used to construct the intent graph (Section~\ref{sec:Gcreate}) and the rest of the GCN inference according to Section~\ref{sec:flexer}.

%We further perform an ablation study to analyze the design of \modelName, reported in Section~\ref{sec:ablation}. In the ablation study, we analyze the effect of the simultaneous multi-label learning and the benefit of the GCN.  

%We further perform an ablation study to analyze the design of \modelName, reported in Section~\ref{sec:ablation}. In the ablation study, we analyze the effect of the simultaneous multi-label learning and the benefit of the GCN.  

\subsubsection{Evaluation Measures}\label{sec:measures}
%\rs{change according to Section~\ref{sec:methodology}}

We evaluate the performance of \er models over a single intent using the standard precision (P), recall (R), and F1 measure (F), as follows.

Let $\mathcal{D} = \{(r_{i}, r_{j}), y_{ij}\}_{(r_i,r_j)\in \candidateset}$ be a labeled set over a candidate record pair set $\candidateset$ for a single intent $\pi$ (see Section~\ref{sec:multi}). The golden standard resolution $\resolution^{*}$ over $\mathcal{D}$, such that $\resolution^{*}\models\theta$, is given by $\resolution^{*} = \{(r_i,r_j)|y_{ij} = 1\}$.  In addition, let $\classifier$ be a resolution creating mapping over $\mathcal{D}$ and $\resolution$ its output, given by $\resolution = \{(r_i,r_j)|\classifier(r_i,r_j) = 1\}$. Precision and recall are computed as follows:

%\ifdefined\TechReport
%\begin{equation}
%P_{\resolution^{*}}(\resolution)=\frac{\mid \resolution\cap \resolution^{*}\mid}{\mid \resolution\mid}, %R_{\resolution^{*}}(\resolution)=\frac{\mid \resolution\cap \resolution^{*}\mid}{\mid \resolution^{*}\mid}
%\end{equation}
%\else
%\begin{scriptsize}
	\begin{equation}
	P_{\resolution^{*}}(\resolution)=\frac{\mid \resolution\cap \resolution^{*}\mid}{\mid \resolution\mid}, R_{\resolution^{*}}(\resolution)=\frac{\mid \resolution\cap \resolution^{*}\mid}{\mid \resolution^{*}\mid}
	\end{equation}
%\end{scriptsize}
%\fi
The F1 measure, $F_{\resolution^{*}}(\resolution)$, is calculated as the harmonic mean of $P_{\resolution^{*}}(\resolution)$ and $R_{\resolution^{*}}(\resolution)$. We use P, R, and F when the context is clear.

The state-of-the-art in \er offers tools that perform extremely well over known benchmarks, \update{especially for datasets that are profiled as easy~\cite{primpeli2020profiling} (see discussion in Section~\ref{sec:benchmarks})}. We augment the standard measures by offering a refined analysis by measuring the \% of baselines error that \modelNameSpace successfully removed. Given two resolutions $\resolution^{\modelNameSpace}$ (outcome of \modelNameSpace) and $\resolution^{baseline}$ (outcome of baseline), and an evaluation measure $V$, where $V\in\{P,R,F\}$, the reduction of residual error is computed as follows:
%\ifdefined\TechReport
%\begin{equation}
%\label{eq:residual_error}
%	E_V(\resolution^{\modelNameSpace},\resolution^{baseline})=100 \cdot \frac{V_{\resolution^{*}}(\resolution^{\modelNameSpace})-V_{\resolution^{*}}(\resolution^{baseline})}{1-V_{\resolution^{*}}(\resolution^{baseline})}
%\end{equation}  
%\else
%\begin{scriptsize}
	\begin{equation}
	\label{eq:residual_error}
	E_V(\resolution^{\modelNameSpace},\resolution^{baseline})=100 \cdot \frac{V_{\resolution^{*}}(\resolution^{\modelNameSpace})-V_{\resolution^{*}}(\resolution^{baseline})}{1-V_{\resolution^{*}}(\resolution^{baseline})}
	\end{equation}  
%\end{scriptsize}
%\fi

To evaluate the quality of models for \problemName, we use the average performance of the single intents per measure. Let $\Pi = \{\pi_{1}, \dots, \pi_{P}\}$ be a set of intents and $V_{\pi}$ be the $V\in\{P,R,F\}$ result with respect to intent $\pi$. The multi intent $V$ performance (MI-V) is computed as
%\ifdefined\TechReport
%\begin{equation}\label{eq:multiC}
% MI\text{-}V = \frac{1}{|\Pi|}\sum_{\pi\in\Pi} V_{\pi}
%\end{equation}
%\else
%\begin{scriptsize} 
	\begin{equation}\label{eq:multiC}
	MI\text{-}V = \frac{1}{|\Pi|}\sum_{\pi\in\Pi} V_{\pi}
	\end{equation}
%\end{scriptsize}
%\fi
In addition, we measure multi label accuracy (Eq.~\ref{eq:accMulti}). Let $\hat{Y}_{ij} = \langle\hat{y}^{1}_{ij}, \hat{y}^{2}_{ij}, \dots, \hat{y}^{P}_{ij}\rangle$ and $Y_{ij} = \langle y^{1}_{ij}, y^{2}_{ij}, \dots, y^{P}_{ij}\rangle$ be the $P$-dimensional (a label for each intent) set of predicted and golden standard intent-based labels of $(r_i, r_j)$, respectively. The multi label accuracy ($MI$-$Acc$) over the candidate set $\candidateset$ is defined as follows.
%\ifdefined\TechReport 
%\begin{equation}\label{eq:accMulti}
%MI\text{-}Acc = \frac{1}{|\candidateset|}\sum_{(r_i, r_j)\in \candidateset}\mathbb{I}(Y_{ij} = \hat{Y}_{ij})
%\end{equation}
%\else
%\begin{scriptsize} 
	\begin{equation}\label{eq:accMulti}
	MI\text{-}Acc = \frac{1}{|\candidateset|}\sum_{(r_i, r_j)\in \candidateset}\mathbb{I}(Y_{ij} = \hat{Y}_{ij})
	\end{equation}
%\end{scriptsize}
%\fi
where $\mathbb{I}(\cdot)$ denotes an indicator function. We report on average $MI\text{-}Acc$ over the tuple pairs in test set. Note that the $MI\text{-}Acc$ is far more strict than $MI\text{-}V$, since it requires the multi intent matcher to be correct over all given intents. %\bg{There is a little problem with the notation here. a regular mean accuracy is a legit measurement and it can be given in equation 7 as well. Maybe we better use a different name than MI for equation 8?}
%\end{sloppypar}
	
Finally, to evaluate the importance of information propagation as a tool to utilize subsumption relationships, we use the measure \emph{preventable error} of intent $\pi$ with respect to resolution $\resolution$ and golden standard resolution $\resolution^{*}$, denoted as $PE_{\pi,\resolution^{*}}\left( \resolution\right) $. This measure is defined as the ratio of false positive predictions that could be prevented by ``listening'' to at least one correct negative prediction of the intents which $\pi$ is subsumed by:
%\ifdefined\TechReport 
%\begin{equation}\label{eq:avoidable_error}
%	PE_{\pi,\resolution^{*}}\left( \resolution\right) = \frac{\left| FP_{\pi,\resolution^{*}}\left( \resolution\right)\right| }{\left| TN_{\pi_{\subseteq},\resolution^{*}}\left( \resolution\right)\right| }
%\end{equation}
%\else
%\begin{scriptsize} 
	\begin{equation}\label{eq:avoidable_error}
	PE_{\pi,\resolution^{*}}\left( \resolution\right) = \frac{\left| FP_{\pi,\resolution^{*}}\left( \resolution\right)\right| }{\left| TN_{\pi_{\subseteq},\resolution^{*}}\left( \resolution\right)\right| }
	\end{equation}
%\end{scriptsize}
%\fi
where $FP_{\pi,\resolution^{*}}\left( \resolution\right)$ is the set of false positive predictions in $\resolution$ with respect to $\pi$, and $TN_{\pi_{\subseteq},\resolution^{*}}\left( \resolution\right)$ is the set of true negative predictions in $\resolution$, such that $\pi_{\subseteq}$ is the \emph{OR} operator defined over the set of intents by which $\pi$ is subsumed.

\subsubsection{Baselines}\label{sec:base}

%The baselines with use to compare with \modelNameSpace are divided into two, first we discuss the ones we compare 

The baselines against which we compare \modelNameSpace solve either a \problemNameSpace or a universal \er problem.

%\vspace{.1cm}
\noindent\textbf{Multi intent baseline:} %Since we are the first to introduce the \problemNameSpace problem, there are no straightforward baselines against which we can compare \modelName. In what follows we suggest 
We use three baselines, as follows. % that represent different approaches to the problem. 
A na\"ive baseline (Na\"ive) assumes that one-size-fits-all and hence, a single solution to the universal \er problem can be applied to multiple intents. The other two baselines were presented in sections~\ref{sec:multi} (In-parallel) and~\ref{sec:prior} (Multi-label). We also report on the single intent performance using these methods. 

%\vspace{.1cm}
\noindent\textbf{Single intent baseline:} We use the state-of-the-art (universal) \er solution DITTO~\cite{li2020deep} (described in Example~\ref{ex:ditto}). Li \emph{et al.} show that DITTO outperforms DeepMatcher~\cite{mudgal2018deep} and follow up works~\cite{fu2019end,kasai2019low,li2020grapher,fu2020hierarchical}. Accordingly, we solely compare \modelNameSpace to DITTO and assume the latter superiority over former models. %Since \modelNameSpace uses DITTO as a part of \modelName, we 
We report the re-produced results of DITTO (see Section~\ref{sec:implementation}) and note that they slightly differ from the ones reported in~\cite{li2020deep}. Specifically, for WDC we modify the training/validation/test sets (see Section~\ref{sec:benchmarks}) and for Walmart-Amazon the differences may be attributed to the removal of two attributes (category and brand) to allow fair predictions over their underlying intents \update{(see Section~\ref{sec:benchmarks}).}
%We separate the report on the {\em equivalence} intent from the report on other intents.} 

\subsection{\protect\problemNameSpace}\label{sec:res_multi}

We first compare \modelNameSpace performance with the baseline methods in solving the \problemNameSpace task. Table~\ref{tab:results_multi} compares the multi intent precision ($MI\text{-}P$), recall ($MI\text{-}R$), and F1 measure ($MI\text{-}F$) (Eq.~\ref{eq:multiC}), multi intent accuracy ($MI\text{-}Acc$, Eq.~\ref{eq:accMulti}) and multi intent reduction of residual error of F1 measure ($MI\text{-}E_F$, Eq.~\ref{eq:residual_error}) of \modelNameSpace and a na\"ive, one-size-fits-all, approach (Na\"ive), the in-parallel approach (In-parallel, Section~\ref{sec:multi}), and multi label learning (Multi-label, Section~\ref{sec:prior}) over the three examined datasets (see Section~\ref{sec:benchmarks}).
%\vskip-.05in
\begin{table}[htpb]
	\centering
	\caption{Multiple intent in terms of $MI\text{-}P$, $MI\text{-}R$, $MI\text{-}F$ (Eq.~\ref{eq:multiC}), $MI\text{-}Acc$ (Eq.~\ref{eq:accMulti}) and $MI\text{-}E_F$ (Eq.~\ref{eq:residual_error}) of \modelNameSpace {\em vs.} baselines.}
	\label{tab:results_multi}
%	\vskip-.05in
	\scalebox{.85}{\begin{tabular}{llccccc}
			\hline
			\textbf{Dataset} & \textbf{Model} & $MI\text{-}P$   & $MI\text{-}R$  & $MI\text{-}F$ & $MI\text{-}Acc$ & $MI\text{-}E_F$ \\\hline
			\multirow{4}{*}{\begin{tabular}[c]{@{}c@{}}AmazonMI\end{tabular}} & Na\"ive      & \update{.831} & \update{.611} & \update{.662} & \update{.769} & -\\
			& In-parallel       & \update{.905} & \update{\textbf{.977}} & \update{.939} & \update{.96} & - \\
			& Multi-label       & \update{.856} & \update{.975} & \update{.907} & \update{.931} & - \\\cdashline{2-7}
			& \modelName    &  \update{\textbf{.951}} & \update{.976} & \update{\textbf{.964}} & \update{\textbf{.977}} & \update{41.0\%}\\\hline
			\multirow{4}{*}{\begin{tabular}[c]{@{}c@{}}Walmart-\\Amazon\end{tabular}} & Na\"ive      & \update{.933} & \update{.282} & \update{.350} & \update{.437}  & -\\
			& In-parallel       & \update{.924} & \update{.918} & \update{.921} & \update{.932}  & -\\
			& Multi-label       & \update{.926} & \update{.919} & \update{.922} & \update{.94} & - \\\cdashline{2-7}
			& \modelName    &  \update{\textbf{.950}} & \update{\textbf{.932}} & \update{\textbf{.94}} & \update{\textbf{.953}} & \update{24.1\%} \\\hline
			\multirow{4}{*}{\begin{tabular}[c]{@{}c@{}}WDC\end{tabular}} & Na\"ive      & \update{.88} & \update{.373} & \update{.459} & \update{.674} & - \\
			& In-parallel       & \update{.876} & \update{.854} & \update{.863} & \update{\textbf{.921}} & - \\
			& Multi-label       & \update{\textbf{.881}} & \update{.836} & \update{.857} & \update{.914} & - \\\cdashline{2-7}
			& \modelName    &  \update{.871} & \update{\textbf{.872}} & \update{\textbf{.871}} & \update{\textbf{.922}} & \update{5.8\%} \\\hline 
	\end{tabular}}
\end{table}	
%\ag{should we also state explicitly the improvement over the naive approach? it is after all, the only one that existed beofre our work. If so, combine with the observation regarding the naive in the next paragraph:} \bg{I don't think so. The naive baseline is just applying the model created for the equivalence intent over the others. Instead I added the sentence about the naive in this paragraph}.
%\vskip-.0725in

\update{As expected, the Na\"ive approach is unsuitable for the \problemNameSpace task, resulting in very low recall values}. Primarily, \modelNameSpace achieves the greatest improvement over the In-parallel baseline results for \emph{AmazonMI} dataset, with average improvement of 2.7\% and 1.8\% in terms of $MI\text{-}F$ and $MI\text{-}Acc$, respectively. These results reflect a reduction of residual error of \update{$41.0\%$ ($MI\text{-}F$) and $42.5\%$ ($MI\text{-}Acc$)}. 
Given that \emph{AmazonMI} organically fits the characteristics of the multiple intent problem, we show that \modelNameSpace is well suited to solve \problemNameSpace by enabling fruitful learning of intent interrelationships. %, leading to an improved prediction.  
As for Walmart-Amazon, \problemNameSpace improves the results of the In-parallel baseline by \update{~2.1\% ($MI\text{-}F$) and ~2.3\% ($MI\text{-}Acc$), namely 24.1\% and ~30.1\%} %in with a corresponding computation of 
reduction of residual error.
\modelNameSpace also achieves an improvement of \update{9.2\% over the In-parallel baseline results for \emph{WDC} dataset in terms of $MI\text{-}F$, albeit with a tiny improvement in terms of $MI\text{-}Acc$}.

\modelNameSpace conjointly predicts accurate labels of different intents, as demonstrated by $MI\text{-}Acc$ improvement. The na\"ive approach offers high precision and low recall, indicating that while a universal solution can provide an accurate resolution, it is also fairly small and incomplete with respect to other interpretations of the dataset.

%\rs{check if true consistently:} The $MI\text{-}F$ improvement is an outcome of an $MI\text{-}P$ boost and a slight $MI\text{-}R$ decline. This may suggest that the inference of \modelNameSpace (using the GCN) can eliminate false-positive predictions of the independent intents. \rs{change the numbers or the example:} For example, \modelNameSpace eliminated $153$ record pairs from the resolution for the \emph{``similar Category-Set''} intent in the AmazonMI dataset, out of which $132$ were correctly classified as $0$ resulting in a 8.3\% precision improvement (see Table~\ref{tab:results_ER_intent}).

\subsection{Single Intent \ER}\label{sec:res_universal}

\modelNameSpace is designed to provide accurate resolutions for the \problemNameSpace task (see Problem~\ref{def:PD}). We next demonstrate that \modelNameSpace can also improve single intent \ernospace, including the equivalence intent representing universal \ernospace~ \update{(Section~\ref{sec:eq_intent}) and other intents (Section~\ref{sec:other_intent}).}
%The results for all other single intents are given in Appendix~\ref{app:res}.

\subsubsection{Equivalence Intent}\label{sec:eq_intent}
\update{Table~\ref{tab:results_eq_intent} provides results in terms of precision ($P$), recall ($R$), F1 measure ($F$), Accuracy ($Acc$) and $E_F$ (see Section~\ref{sec:measures}) for equivalence intent solely, which is the universal \er interpretation.} 

\begin{table}[htpb]
	\centering
	\caption{\update{Equivalence intent results in terms of precision (P), recall (R), F1 (F), Acc and $E_F$.}}
	\label{tab:results_eq_intent}
		\scalebox{.8}{\begin{tabular}{llccccc}
			\hline
			\textbf{Dataset} & \textbf{Model} & $P$   & $R$  & $F$ & $Acc$ & $E_F$ \\\hline
			\multirow{3}{*}{\begin{tabular}[c]{@{}c@{}}AmazonMI\end{tabular}}
			& In-parallel       & \update{.829} & \update{\textbf{.991}} & \update{.901} & \update{.960} & \update{-} \\
			& Multi-label       & \update{.921} & \update{.905} & \update{.912} & \update{.969} & \update{-} \\\cdashline{2-7}
			& \modelName    & \update{\textbf{.933}} & \update{.985} & \update{\textbf{.958}} & \update{\textbf{.985}} & \update{57.6 \%} \\\hline
			\multirow{3}{*}{\begin{tabular}[c]{@{}c@{}}Walmart-\\Amazon\end{tabular}}
			& In-parallel       & \update{.852} & \update{\textbf{.812}} & \update{.831} & \update{.969}  & \update{-} \\
			& Multi-label       & \update{.854} & \update{.772} & \update{.810} & \update{.966} & \update{-} \\\cdashline{2-7}
			& \modelName    &  \update{\textbf{.903}} & \update{.792} & \update{\textbf{.844}} & \update{\textbf{.985}} & \update{7.7 \%}  \\\hline
			\multirow{3}{*}{\begin{tabular}[c]{@{}c@{}}WDC\end{tabular}}
			& In-parallel       & \update{.786} & \update{.745} & \update{.761} & \update{.948} & \update{-} \\
			& Multi-label       & \update{\textbf{.808}} & \update{.713} & \update{.757} & \update{.948} & \update{-} \\\cdashline{2-7}
			& \modelName    &  \update{.775} & \update{\textbf{.788}} & \update{\textbf{.782}} & \update{\textbf{.950}} & \update{8.8 \%} \\\hline 
	\end{tabular}}
\end{table}

%We compare precision ($P$), recall ($R$), and F1 measure ($F$) (see Section~\ref{sec:measures}) of 
%\modelNameSpace to DITTO (In-parallel) and Multi-label (see Section~\ref{sec:base}) for each single intent over the four examined datasets (see Section~\ref{sec:benchmarks}). The top comparison for each dataset presents the equivalence intent, which is the universal \er interpretation, where the proportional improvement of \modelNameSpace over DITTO is given in parenthesis for each dataset. 

\update{Our experiments indicate that \modelNameSpace offers an improved performance of the equivalence intent (universal \ernospace). Compared to the state-of-the-art (DITTO), \modelNameSpace demonstrates a substantial increase of 6.3\% for the \emph{AmazonMI} dataset in terms of $F$ measure (reduction of residual error of 57.6\%). \modelNameSpace also shows an improvement of 1.6\% and 2.8\% ($F$ measure) compared to DITTO for the \emph{Walmart-Amazon} and \emph{WDC} dataset, respectively. These results indicate that with information of multiple intents annotations (\emph{e.g.}, derived from available meta data) \modelNameSpace is an effective solution for the universal \ernospace~problem.}

\subsubsection{Other Intents}\label{sec:other_intent}
\update{Table~\ref{tab:results_ER_intent} provides results in terms of precision ($P$), recall ($R$), F1 measure ($F$), Accuracy ($Acc$) and $E_F$ (see Section~\ref{sec:measures}) for all datasets and their intents (except the equivalence intent that is reported in Table~\ref{tab:results_eq_intent}). 
We compare \modelNameSpace with the same baselines as in Section~\ref{sec:eq_intent} for each single intent.}
%Due to space consideration we report only on the results of AmazonMI, and refer the reader to the technical report~\cite{GitTechReport} for the full results.} 	

\ifdefined\TechReport
	\begin{table}[htpb]
		\centering
		\caption{Single intent results \update{except equivalence} in terms of precision (P), recall (R), F1 (F), Acc and $E_F$.}
		\label{tab:results_ER_intent}
		\scalebox{.71}{\begin{tabular}{lllccccc}
				\hline
				\textbf{Dataset} & \textbf{Intent} & \textbf{Model} & $P$   & $R$  & $F$ & Acc & $E_F$ \\\hline
				\multirow{12}{*}{\begin{tabular}[c]{@{}c@{}}AmazonMI\end{tabular}}
				& \multirow{3}{*}{\begin{tabular}[c]{@{}c@{}}Brand\end{tabular}}  & DITTO (In-parallel)  & \update{.926} & \update{.978} & \update{.951}   & \update{.981} & \update{-} \\
				&& Multi-label  & \update{.856} & \update{\textbf{.993}} & \update{.919}   & \update{.965} & \update{-} \\\cdashline{3-8}
				& & \modelName    & \update{\textbf{.934}} & \update{.979} & \update{\textbf{.956}} & \update{\textbf{.982}} & \update{10.2\%} \\\cline{2-8}
				& \multirow{3}{*}{\begin{tabular}[c]{@{}c@{}}Set-Cat.\end{tabular}}  & DITTO (In-parallel)  & \update{.912} & \update{.977} & \update{.944} & \update{.944} & \update{-}  \\
				&& Multi-label  & \update{.908} & \update{\textbf{.990}} & \update{.947} & \update{.947} & \update{-} \\\cdashline{3-8}
				& & \modelName     & \update{\textbf{.968}} & \update{.976} & \update{\textbf{.972}} & \update{\textbf{.973}} & \update{50.0\%} \\\cline{2-8}
				& \multirow{3}{*}{\begin{tabular}[c]{@{}c@{}}Main-Cat.\end{tabular}}  & DITTO (In-parallel)  & \update{.979} & \update{.989} & \update{.984} & \update{.978} & \update{-}  \\
				&& Multi-label  & \update{.945} & \update{\textbf{.993}} & \update{.969} & \update{.957} & \update{-} \\\cdashline{3-8}
				& & \modelName     & \update{\textbf{.988}} & \update{.987} & \update{\textbf{.988}} & \update{\textbf{.983}} & \update{25.0\%} \\\cline{2-8}
				& \multirow{3}{*}{\begin{tabular}[c]{@{}c@{}}Main-Cat. +\\ Set-Cat.\end{tabular}}  & DITTO (In-parallel)  & \update{.881} & \update{.948} & \update{.913} & \update{.937} & \update{-}    \\
				&& Multi-label  & \update{.65} & \update{\textbf{.993}} & \update{.786} & \update{.815} & \update{-}  \\\cdashline{3-8}
				& & \modelName    & \update{\textbf{.932}}  & \update{.955}  & \update{\textbf{.944}} & \update{\textbf{.961}} & \update{35.6\%} \\\cline{2-6} 
				\hline 
				\multirow{9}{*}{\begin{tabular}[c]{@{}c@{}}Walmart-Amazon\end{tabular}} & \multirow{3}{*}{\begin{tabular}[c]{@{}c@{}}Brand\end{tabular}}  & DITTO (In-parallel)  & \update{.977} & \update{.964} & \update{.971} & \update{.955} & \update{-}   \\
				&& Multi-label  & \update{.970} & \update{.976} & \update{.973} & \update{.959} & \update{-}   \\\cdashline{3-8}
				& & \modelName    & \update{\textbf{.986}}  & \update{\textbf{.990}} & \update{\textbf{.988}} & \update{\textbf{.973}} & \update{43.6\%} \\\cline{2-8}
				& \multirow{3}{*}{\begin{tabular}[c]{@{}c@{}}Main-Cat.\end{tabular}}  & DITTO (In-parallel)  & \update{.921} & \update{.931} & \update{.926} & \update{.881} & \update{-}   \\
				&& Multi-label  & \update{.927} & \update{.952} & \update{.939} & \update{.901} & \update{-}  \\\cdashline{3-8}
				& & \modelName    & \update{\textbf{.942}}  & \update{\textbf{.959}} & \update{\textbf{.950}} & \update{\textbf{.911}} & \update{32.5\%}  \\\cline{2-8}
				& \multirow{3}{*}{\begin{tabular}[c]{@{}c@{}}General-Cat.\end{tabular}}  & DITTO (In-parallel)  & \update{0.948} & \update{.968} & \update{.957} & \update{.922} & \update{-}   \\
				&& Multi-label  & \update{.954} & \update{.976} & \update{.965} & \update{.936} & \update{-}  \\\cdashline{3-8}
				& & \modelName    & \update{\textbf{.967}}  & \update{\textbf{.987}} & \update{\textbf{.977}} & \update{\textbf{945}} & \update{46.5\%} \\\cline{2-8}
				\hline 
				\multirow{6}{*}{\begin{tabular}[c]{@{}c@{}}WDC\end{tabular}} & \multirow{3}{*}{\begin{tabular}[c]{@{}c@{}}Category\end{tabular}}  & DITTO (In-parallel)  & \update{\textbf{.939}} & \update{.880} & \update{.909} & \update{.923} & \update{-}   \\
				&& Multi-label  & \update{.934} & \update{.889} & \update{\textbf{.911}} & \update{\textbf{.924}} & \update{-}   \\\cdashline{3-8}
				& & \modelName    & \update{.932}  & \update{\textbf{.89}} & \update{\textbf{.911}} & \update{.923} & \update{1.0\%} \\\cline{2-8}
				& \multirow{3}{*}{\begin{tabular}[c]{@{}c@{}}General-Cat.\end{tabular}}  & DITTO (In-parallel)  & \update{\textbf{.904}} & \update{.937} & \update{.92} & \update{\textbf{.891}} & \update{-}   \\
				&& Multi-label  & \update{.902} & \update{.905} & \update{.904} & \update{.870} & \update{-}   \\\cdashline{3-8}
				& & \modelName    & \update{.900}  & \update{\textbf{.943}} & \update{\textbf{.921}} & \update{\textbf{.891}} & \update{1.0\%} \\\cline{2-8}
				\hline 
		\end{tabular}}
	\end{table}
\else
	\begin{table}[htpb]
		\centering
		\caption{Single intent results \update{except equivalence} in terms of precision (P), recall (R), F1 (F), Acc and $E_F$.}
		\label{tab:results_ER_intent}
		\scalebox{.71}{\begin{tabular}{lllccccc}
				\hline
				\textbf{Dataset} & \textbf{Intent} & \textbf{Model} & $P$   & $R$  & $F$ & Acc & $E_F$ \\\hline
				\multirow{12}{*}{\begin{tabular}[c]{@{}c@{}}AmazonMI\end{tabular}}
				& \multirow{3}{*}{\begin{tabular}[c]{@{}c@{}}Brand\end{tabular}}  & DITTO (In-parallel)  & \update{.926} & \update{.978} & \update{.951}   & \update{.981} & \update{-} \\
				&& Multi-label  & \update{.856} & \update{\textbf{.993}} & \update{.919}   & \update{.965} & \update{-} \\\cdashline{3-8}
				& & \modelName    & \update{\textbf{.934}} & \update{.979} & \update{\textbf{.956}} & \update{\textbf{.982}} & \update{10.2\%} \\\cline{2-8}
				& \multirow{3}{*}{\begin{tabular}[c]{@{}c@{}}Set-Cat.\end{tabular}}  & DITTO (In-parallel)  & \update{.912} & \update{.977} & \update{.944} & \update{.944} & \update{-}  \\
				&& Multi-label  & \update{.908} & \update{\textbf{.990}} & \update{.947} & \update{.947} & \update{-} \\\cdashline{3-8}
				& & \modelName     & \update{\textbf{.968}} & \update{.976} & \update{\textbf{.972}} & \update{\textbf{.973}} & \update{50.0\%} \\\cline{2-8}
				& \multirow{3}{*}{\begin{tabular}[c]{@{}c@{}}Main-Cat.\end{tabular}}  & DITTO (In-parallel)  & \update{.979} & \update{.989} & \update{.984} & \update{.978} & \update{-}  \\
				&& Multi-label  & \update{.945} & \update{\textbf{.993}} & \update{.969} & \update{.957} & \update{-} \\\cdashline{3-8}
				& & \modelName     & \update{\textbf{.988}} & \update{.987} & \update{\textbf{.988}} & \update{\textbf{.983}} & \update{25.0\%} \\\cline{2-8}
				& \multirow{3}{*}{\begin{tabular}[c]{@{}c@{}}Main-Cat. +\\ Set-Cat.\end{tabular}}  & DITTO (In-parallel)  & \update{.881} & \update{.948} & \update{.913} & \update{.937} & \update{-}    \\
				&& Multi-label  & \update{.65} & \update{\textbf{.993}} & \update{.786} & \update{.815} & \update{-}  \\\cdashline{3-8}
				& & \modelName    & \update{\textbf{.932}}  & \update{.955}  & \update{\textbf{.944}} & \update{\textbf{.961}} & \update{35.6\%} \\\cline{2-6} 
				\hline 
		\end{tabular}}
	\end{table}
\fi

\update{The results for \emph{AmazonMI} demonstrates %provides the following intriguing insight into 
	the ability of \modelNameSpace to benefit from intent interrelationships.
The improvement achieved by \modelNameSpace over the intents of \emph{%equivalence, 
	set-category} and \emph{set-category+main-category} is notably higher than the other two intents. The common thread between the %three 
two involves the subsumption relationship (Definition~\ref{def:subsum}). 
\emph{set-category} and \emph{set-category+main-category} are subsumed by \emph{main-category}. %, and \emph{equivalence} is subsumed by all other intents. 
This implies that \modelNameSpace manages to propagates information among levels of granularity.} 

\subsection{Inter-Layer Edge Analysis}\label{sec:inter_layer_analysis}
\update{In this subsection we examine the contribution of intents in solving universal \ernospace, and show that learning interconnections can help in avoiding \emph{preventable error} (Eq.~\ref{eq:avoidable_error}).}

\subsubsection{Intent Interrelationships}\label{sec:Interrelationships}
\begin{figure*}[t]
	\centering
	\begin{subfigure}[b]{0.28\textwidth}
		\centering
		\includegraphics[width=\textwidth]{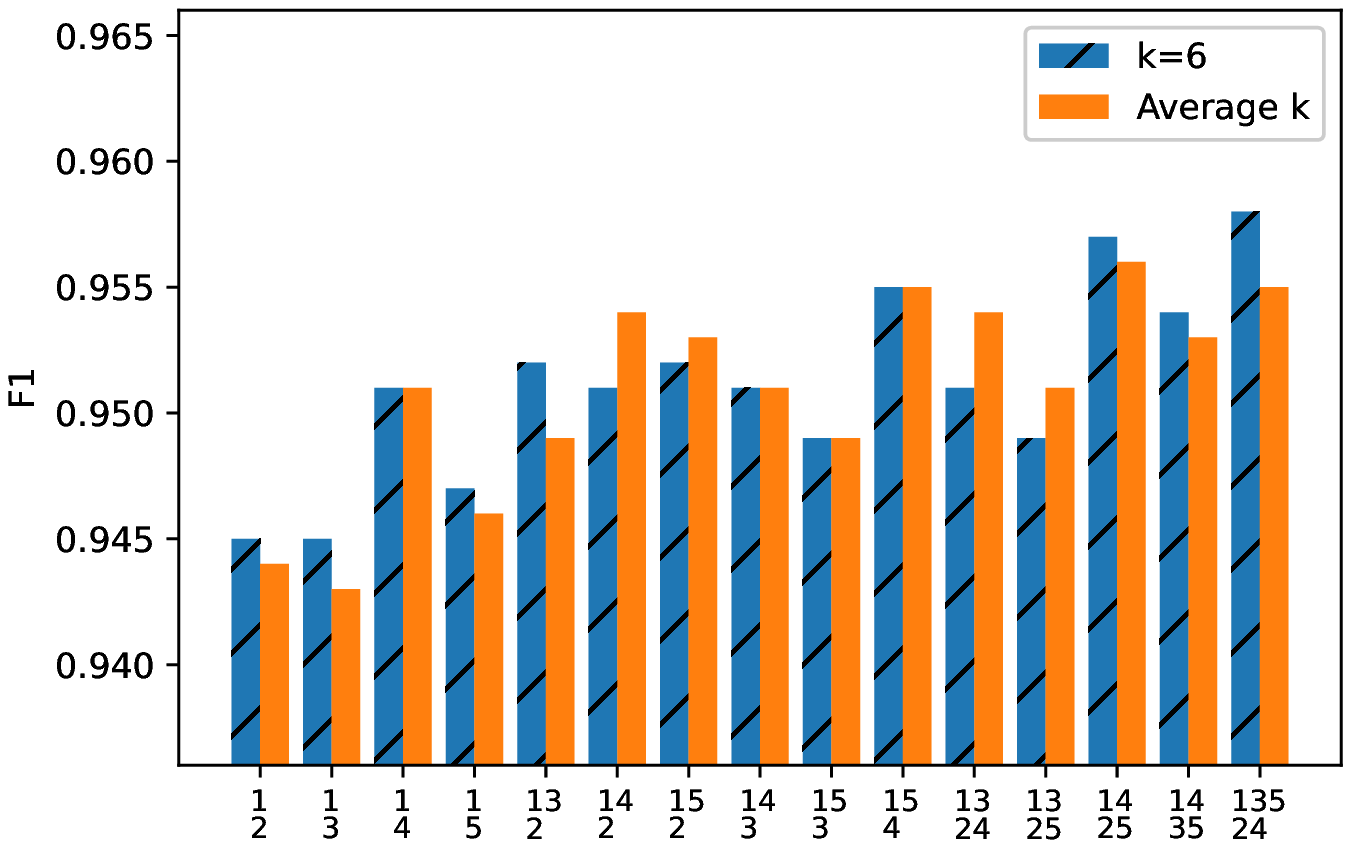}
%		\vskip -0.1in
		\caption{AmazonMI}
		\label{fig:AmazonMI_intent_ablation}
	\end{subfigure}
	\hfill
	\begin{subfigure}[b]{0.28\textwidth}
		\centering
		\includegraphics[width=\textwidth]{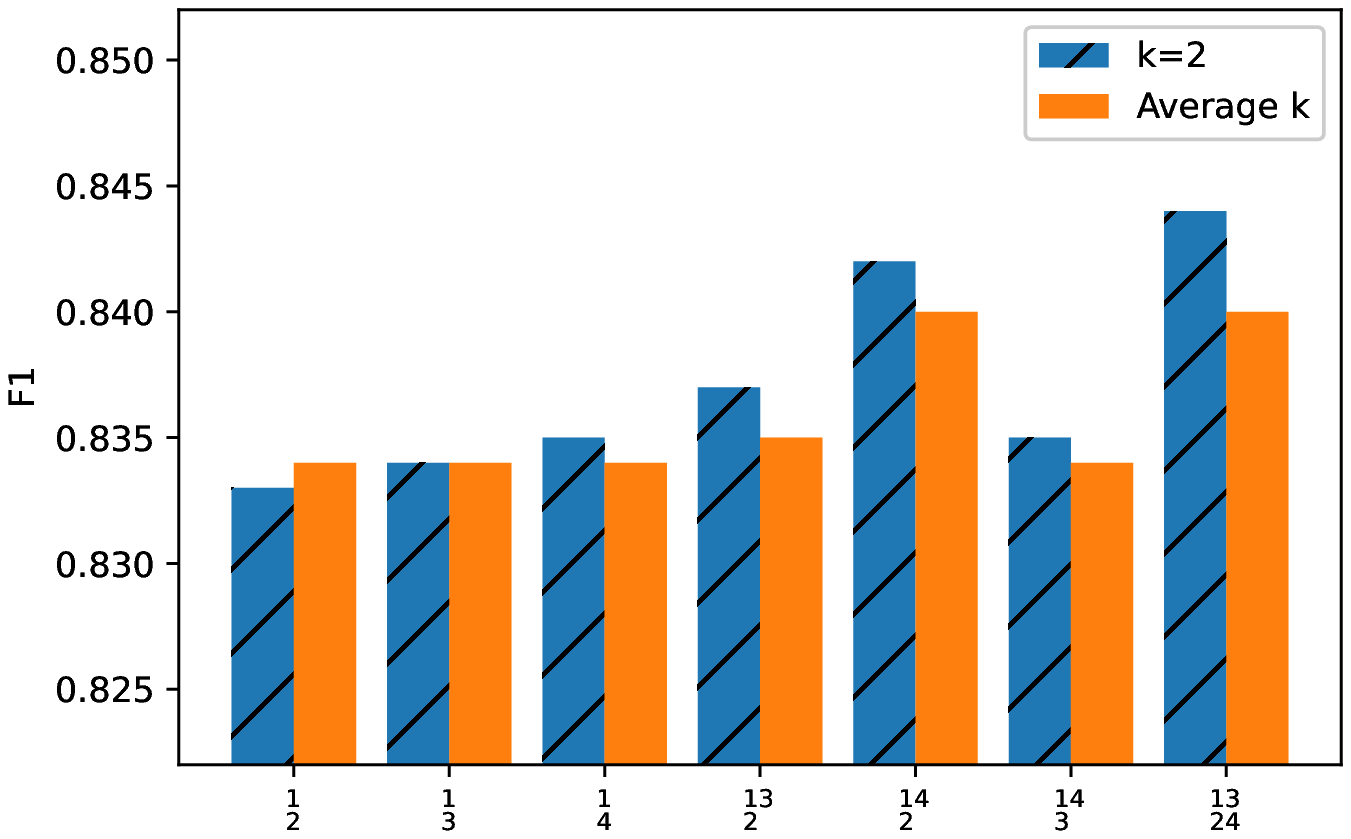}
%		\vskip -0.1in
		\caption{Walmart-Amazon}
		\label{fig:WA_intent_ablation}
	\end{subfigure}
	\hfill
	\begin{subfigure}[b]{0.28\textwidth}
		\centering
		\includegraphics[width=\textwidth]{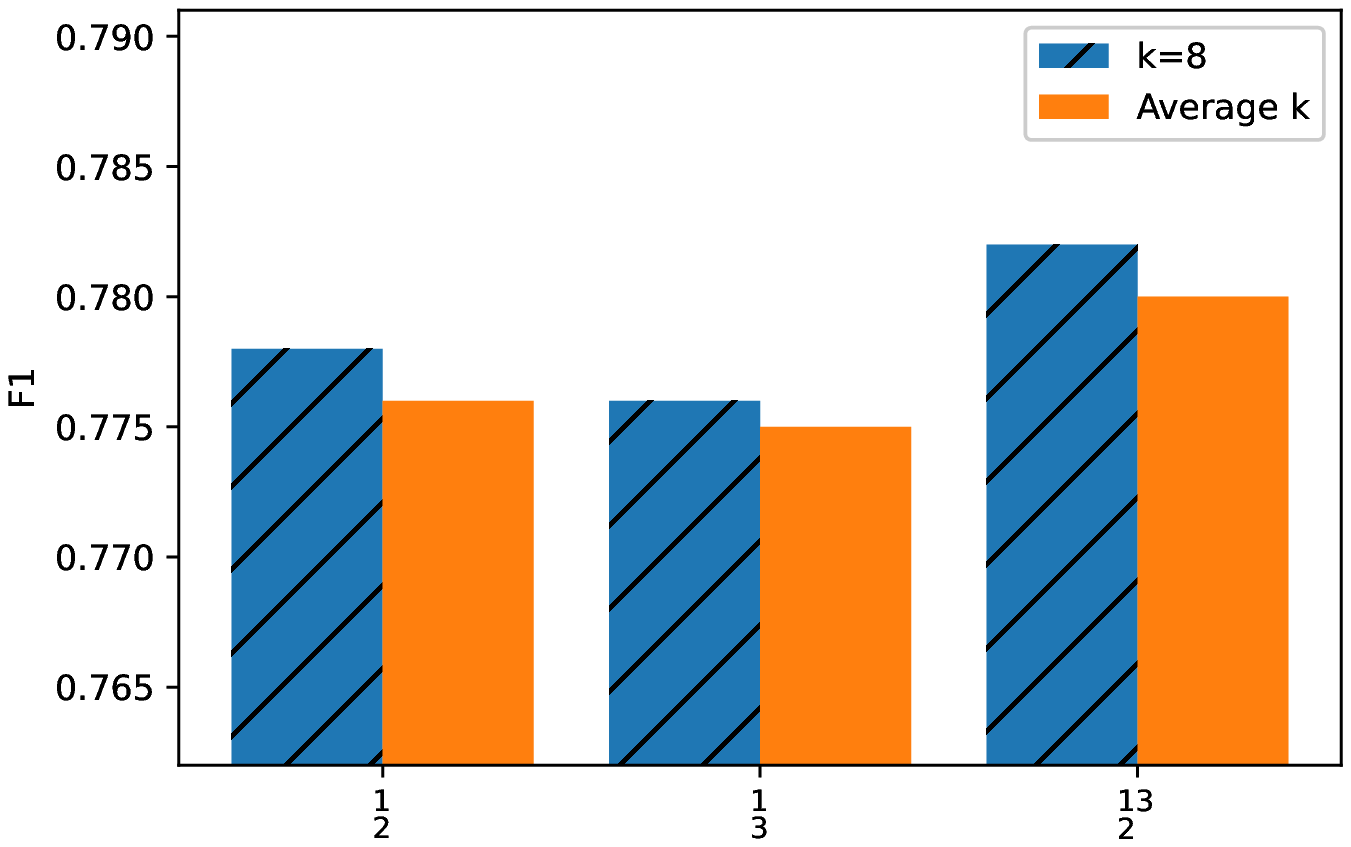}
%		\vskip -0.1in
		\caption{WDC}
		\label{fig:WDC_intent_ablation}
	\end{subfigure}
%	\vskip -0.1in
	\caption{\update{Performance (in terms of F1 for the equivalence intent) as a function of the intents that were used in building the multiplex graph. The numbers under each pair of bars refer to the identifiers of the intents, as they appear in Table~\ref{tab:Datasetsapp}.}}
	\label{fig:Interrelationships}
\end{figure*}

\update{For each dataset we found the parameters ($h_1$, $k$ and number of GNN layers) leading to the best average F1 value achieved over the universal \ernospace ~(equivalence intent), and fixed them. For each combination of these parameters, we generated the multiplex graph with every subset of the complete intent set which contains the equivalence intent. In addition, we also computed the average result over all possible k values ($k \in \left\lbrace 0,2,4,6,8,10\right\rbrace $) when $h_1$ remains fixed.}
	
\update{Figure~\ref{fig:Interrelationships} demonstrates the effectiveness of utilizing intent interrelationships, suggesting that in most cases the more is better. The numbers in the figure denote the intents according to their numbering in Table~\ref{tab:Datasetsapp}. For all datasets, the best result is achieved for the entire intent set, suggesting that \modelName ~benefits from additional information (more intents in our case) while solving the standard universal \ernospace ~task.} %\bg{I would like to rephrase this sentence. Talking about monotonicity is a bit misleading. I just want to say that in general the trend is "more is better"}The model performance is monotonic in the number of intents involved in the model, such that a model built upon a certain number of intents almost always outperforms a model created over a smaller number of intents. This statement holds true both when computed only for the best $k$ and when averaged over the entire set of $k$ values.}

\subsubsection{Intents Ablation Study}\label{sec:intent_ablation}
\begin{sloppypar}
To measure the benefit of relying on intent interelationships, we use the measure of preventable error (Eq.~\ref{eq:avoidable_error}), Focusing on \emph{AmazonMI} and using the hyperparameters which lead to the best F1 score over the equivalence intent. Figure~\ref{fig:PE_graph} demonstrates the gaps between \modelName and the in-parallel baseline in terms of this measure, as \update{$PE_{Eq,\resolution^{*}}\left( \resolution^{\modelNameSpace}\right) = 7.97 \cdot 10^{-4} $}, while $PE_{Eq,\resolution^{*}}\left( \resolution^{baseline}\right) = 15.89 \cdot 10^{-3}$, which is approximately $20$ times more. This conclusion holds also for the \emph{set-category} intent with \update{$PE_{Set\mbox{-}Cat,\resolution^{*}}\left( \resolution^{\modelNameSpace}\right) = 2.0 \cdot 10^{-3} $}, $PE_{Set\mbox{-}Cat,\resolution^{*}}\left( \resolution^{baseline}\right) = 6.3 \cdot 10^{-2}$, and for the \emph{main-category+set-category} intent with \update{$PE_{Main\mbox{-}Cat+Set\mbox{-}Cat,\resolution^{*}}\left( \resolution^{\modelNameSpace}\right) = 2.0 \cdot 10^{-3} $}, $PE_{Main\mbox{-}Cat+Set\mbox{-}Cat,\resolution^{*}}\left( \resolution^{baseline}\right) = 2.1 \cdot 10^{-2}$. In both cases, we see an order of magnitude more preventable error.  
\end{sloppypar}

Recall that real-world meaning of intents is unknown to the model. Therefore, relationships between intents cannot be directly derived. The above analysis reveals that \modelNameSpace offers an effective approach to learn those dependencies via a message propagation mechanism, significantly reducing the rate of preventable error.  

\begin{figure}[htpb]
	\centering
	\includegraphics[width=0.625\columnwidth]{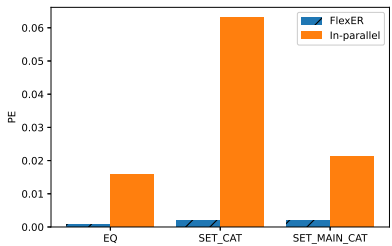}
	\vskip -0.1in
	\caption{Preventable Error: \modelNameSpace {\em vs.} in-parallel} 
	\label{fig:PE_graph}
\end{figure}

\subsection{Intra-Layer Edge Analysis}\label{sec:intra_layer_analysis}
%\begin{figure*}[htpb]
%	\centering
%	\begin{subfigure}[b]{0.28\textwidth}
%		\centering
%		\includegraphics[width=\textwidth]{figs/AmazonMI_k_ablation.eps}
%		\caption{AmazonMI}
%		\label{fig:AmazonMI_k_ablation}
%	\end{subfigure}
%	\hfill
%	\begin{subfigure}[b]{0.28\textwidth}
%		\centering
%		\includegraphics[width=\textwidth]{figs/WA_k_ablation.eps}
%		\caption{Walmart-Amazon}
%		\label{fig:WA_k_ablation}
%	\end{subfigure}
%	\hfill
%	\begin{subfigure}[b]{0.28\textwidth}
%		\centering
%		\includegraphics[width=\textwidth]{figs/WDC_k_ablation.eps}
%		\caption{WDC}
%		\label{fig:WDC_k_ablation}
%	\end{subfigure}
%	\caption{\update{Performance (in terms of F1 for the equivalence intent) as a function of $k$ (number of intra-layer edges connected to each node).}}
%	\label{fig:k_ablation}
%\end{figure*}
\update{%The intuition behind the design choice of adding intra-layer edges comes from the K nearest neighbor algorithm, where vector-space similarity is interpreted as agreement between samples~\cite{bhatia2010survey}. We now empirically show that adding intra-layer edges improves \modelName's performance. 
%	Average F1 value over the universal \ernospace ~(equivalence intent) was calculated over the entire set of available intents, and $2)$ over the entire set of intent combinations containing the equivalence intent.
%\ag{the following is unclear:} building the model over the entire set of intent and average performance over the entire set of intent combinations containing the equivalence intent. 
%As described in Section ~\ref{sec:implementation}, we 
We tested \modelName's performance using $k \in \left\lbrace 0,2,4,6,8,10\right\rbrace$, and alongside the case of $k=0$ (no intra-layer edges) we report the average F1 value over all positive $k$ values for the universal \ernospace ~(equivalence intent). For these experiments we built the multiplex graph over the entire set of available intents.}

\begin{table}[t]
	\centering
	\caption{Analysis of $k$ value in terms of F1.}
%	\vskip -0.1in
	\scalebox{.8}{\begin{tabular}{|l|c|c|c|c|}
			\hline
			& AmazonMI & Walmart-Amazon  & WDC \\\hline
			\multirow{1}{*}{\begin{tabular}[c]{@{}c@{}}\textbf{k=0}\end{tabular}}
			& .951  & .833 & .772 \\\hline
			\multirow{1}{*}{\begin{tabular}[c]{@{}c@{}}\textbf{k>0}\end{tabular}}
			& .955(+.42\%)  & .838(+.60\%) & .777(+.65\%) \\\hline
%			\multirow{1}{*}{\begin{tabular}[c]{@{}c@{}}\textbf{+(\%)}\end{tabular}}
%			& .42  & .60 & .65 \\\hline
	\end{tabular}}
	\label{tab:k_ablation}
%	\vskip -0.15in
\end{table}

\update{The experiment results are shown in Table~\ref{tab:k_ablation}, indicating that adding intra-layer edges is indeed effective. In both cases, the results obtained when $k=0$ are inferior to $k>0$. 
%The experiment do not reveal any significant difference between the positive $k$ values though (\bg{It does not appear here right now since we aggregated the positive results. Maybe say something like this is the reason for the aggregation?}). 
Also, no clear dominance of a specific $k$ value is evident and the choice of $k$ remains a design choice that depends on the dataset. To summarize, the experiments strengthen the results presented in Section~\ref{sec:intent_ablation}, as the results obtained by using the entire set of available intents are constantly better than the average values over all intent combinations containing the equivalence intent.}

\subsection{Run-Time Analysis}\label{sec:time_complexity}
\update{\modelName~is built on top of intent-based representation, created by applying several DITTO models, one per intent. 
With adequate resources, those DITTO models can be trained simultaneously, such that considering multiple intent does not require extra time compared to an ordinary training of DITTO.} 

\update{There are two modes of creating the multiplex graph. The first is performed exactly once for an entire hyperparamter set, including the nearest neighbors (NN) computation (Section~\ref{sec:Gintraedges}). The second directly loads an available nearest neighbors dictionary, which can be done once the former mode is executed.}

\update{The run-times of \modelName~ over the three datasets (in seconds) are presented in Table~\ref{tab:running_time}.
The nearest neighbors computation is orthogonal to the selection of hyperparameters. Hence, we separate the report of this calculation from the model training and testing. The reported run-times refer to the nearest neighbors computation of train, validation, and test dataset combined. It is not surprising that the size of the datasets directly affect this computation, as WDC requires significantly more time than the other two, while AmazonMI requires more time than Walmart-Amazon. Note that in our experiments we use only the exhaustive version of nearest neighbors computation, although~\emph{Faiss}~\cite{johnson2019billion} offers multiple heuristics that can reduce the computational effort.}

\update{We also report on the run-time of the training and testing phase (over 150 epochs). While the hyperparameter selection of $h_1$ and $k$ does not drastically affect the run-times, we still report on average results over all examined combinations of these two. $L$, the third hyperparameter, denotes the number of layers (2 or 3) in the GNN, which actually affects the training time. The results show that the burden of training and testing \modelName is negligible comparing to the preparatory phase of training DITTO model (approximately two orders of magnitude less). To conclude, once nearest neighbors dictionary is created, \modelName~can be easily optimized over large set of hyperparameters due to its fast training time.}

\begin{table}[t]
	\centering
	\caption{Average run-time of \modelName~(in seconds).}
%	\vskip -0.1in
	\scalebox{.8}{\begin{tabular}{|l|c|c|c|}
			\hline
			& AmazonMI & Walmart-Amazon  & WDC \\\hline
			\multirow{1}{*}{\begin{tabular}[c]{@{}c@{}}\textbf{NN Computation}\end{tabular}}
			& 398.6  & 139.5 & 954.5 \\\hline
			\multirow{1}{*}{\begin{tabular}[c]{@{}c@{}}\textbf{Training+Testing(2L)}\end{tabular}}
			& 11.4  & 8.1 & 6.7 \\
			\multirow{1}{*}{\begin{tabular}[c]{@{}c@{}}\textbf{Training+Testing(3L)}\end{tabular}}
			& 16.7  & 11.9 & 9.0 \\\hline
	\end{tabular}}
	\label{tab:running_time}
%	\vskip -0.25in
\end{table}

%\input{results3}
%\input{results4}

%\input{results}

%\newpage
\section{Related Work}\label{sec:related}

%\ag{perhaps add here:} The idea of creating several diffrent views over a dataset is also known in the literature as 
%\emph{materialized views}~\cite{gupta1995maintenance}.

Over the years, multiple solutions were suggested to tackle end-to-end \er~\cite{cohen2002learning,papadakis2020three} and its various steps (\emph{e.g.,} blocking~\cite{papadakis2016comparative,papadakis2020blocking} and matching~\cite{singh2017synthesizing,li2020deep}). Overall, the main goal of \er is to find equivalent record pairs, assuming that the more commonality they share the more similar they are~\cite{lin1998information}. This observation has led to multiple similarity measure methods~\cite{levenshtein1966binary,jaro1989advances,jaro1995probabilistic} that accurately estimate equality~\cite{wang2011entity}. More recent works aimed to improve such estimation by fusing similarity measures~\cite{bilenko2003adaptive,konda2016magellan}. Others aimed to extract rules from data for matching tuples~\cite{singla2006entity,singh2017synthesizing}, \update{and recently even the usage of meta-learning techniques~\cite{miao2021rotom} was proposed.} All of these methods address the universal \er problem (Section~\ref{sec:ER}), which, in the context of this paper, means resolving a single equivalence intent. We define and address the problem of \problemName, extending universal \er to support multiple resolution intents (including the equivalence intent). \problemNameSpace is motivated by the need to support multiple entity interpretations in downstream applications. Our proposed solution, \modelName, is shown empirically to be backward compatible in that it offers an improved solution to the universal resolution intent as well.

Similar to other data integration tasks~\cite{cappuzzo2020creating,thirumuruganathan2020data}, deep learning is used to tackle \er problems. Ebraheem {\em et al}.~\cite{joty2018distributed} were the first to use neural networks for \ernospace, utilizing the contextual similarity of attribute values. Mudgal {\em et al}.~\cite{mudgal2018deep} addressed the variation of entity matching by introducing a design space for the use of deep learning, which was later extended using multi-perspective matching~\cite{fu2019end}, transfer learning~\cite{kasai2019low,zhao2019auto} and hierarchical network~\cite{fu2020hierarchical}. %Closest to our approach is the 
Li {\em et al.}~\cite{li2020grapher} use graph convolutional networks for the \er task, albeit using a different representation of tuple pairs than ours. %, see Section~\ref{sec:Gcreate}. 
%which is expected, given the increased utilization of contextual embeddings in data cleaning and integration at large~\cite{cappuzzo2020creating,thirumuruganathan2020data}.

%The use of contextualized word embeddings using pre-trained language models is beneficial in entity resolution (see~\cite{brunner2020entity,li2020deep}), which is expected given its increasing utilization in data cleaning and integration~\cite{cappuzzo2020creating,thirumuruganathan2020data}.   
%%Recently,  were also introduced for the task of entity matching/resolution~\cite{brunner2020entity,li2020deep}, the latter
%%DITTO~\cite{li2020deep} is presented in Example~\ref{ex:ditto}. 
%Recent works~\cite{li2021improving,peeters2021dual} expanded the usage of pre-trained language models by suggesting new variations of architectures, utilizing the standard BERT~\cite{devlin:2018bert} to form an improved version and allowing the network to train over combination of more than a single loss function. 
%\modelNameSpace uses learning-based matchers to address a new \er problem, that of multiple intents (including equivalence), supported by graph convolutional networks, which also improves on the universal \er (equivalence intent) problem.

The use of contextualized word embeddings based on pre-trained language models %(PTLMs) 
is beneficial in entity resolution \update{(see~\cite{li2021deep, brunner2020entity,li2020deep}). Such models, and specifically BERT~\cite{devlin:2018bert}, can cope with semantic heterogeneity in language and structural diversity in data instances. Studies~\cite{li2020deep,brunner2020entity} established the superiority of this approach over former methods by serializing input data to form a long string separated by artificial separator tokens, and fine-tuning a language model on top of the desired dataset. DITTO~\cite{li2020deep} enhances performance further using data augmentation, text summarization, and domain knowledge injection. DITTO is used in our work as a baseline, producing record pair representations (see Example~\ref{ex:ditto}).} %Li {\em et al.}~\cite{li2021deep} thoroughly surveys DITTO's components and explores additional directions beyond entity matching of which DITTO can be applied for.}
	
Recent works~\cite{li2021improving,peeters2021dual} %expanded the usage of PTLMs by suggesting 
suggest new variations of BERT-based structure architectures for entity resolution, allowing the network to train over a combination of more than a single loss function. 
\update{Peeters {\em et al.}~\cite{peeters2020intermediate} focus on product matching, a prevalent scenario of entity matching. They show that an intermediate training step over product datasets improves the effectiveness of BERT-based models to the task. The usage of pre-trained models is also extended to blocking~\cite{thirumuruganathan2021deep}}. %the preparatory phase of entity resolution (see Section~\ref{sec:ER}), where Thirumuruganathan {\em et al.}~\cite{thirumuruganathan2021deep} offer various deep learning solutions, including BERT-based ones, to tackle this problem.}
\modelNameSpace uses pre-trained models to address a new \er problem, that of multiple intents (including equivalence), supported by graph convolutional networks, which also offers an improved solution to the universal \er (equivalence intent) problem.

Several related works~\cite{andritsos2006clean,soliman2007top,DONG2009} propose to examine multiple \er problems (possible worlds) by attaching probabilities to resolutions/tuple pairs, to solve the universal \er problem.  Ensemble \er approaches~\cite{zhao2005entity,chen2009exploiting,jurek2017novel,meduri2020comprehensive} combine multiple individual solutions to provide a more accurate \textbf{single} solution. \modelNameSpace solves a \problemNameSpace problem, where different interpretations to the notion of a real-world entity lead to various intents with different solutions.

%Multiple intents can be found in various NLP problems. 
Recent works concentrate on leveraging external structure-based knowledge %, which can be seen as analogous to an intent system, 
for extracting information. Karamanolakis {\em et al.}~\cite{karamanolakis2020txtract} focus on knowledge extraction using taxonomy with the objective of extracting category-specific attribute values based on the hierarchical structure of categories. Yu {\em et al.}~\cite{yu2020steam} target expansion of taxonomy terms using natural supervision of existing taxonomies. Taxonomy is also shown to be valuable in the \er pipeline, as part of the blocking phase~\cite{wang2015semantic}, introducing a semantic similarity measure that relies on taxonomy trees. These works are distinct from ours. Whereas taxonomy-based frameworks start from a known %are typically aware of the structured nature of the problem 
taxonomy and make use of it, we do not assume intents structure is unknown. Rather, %known information (
labels of training data are encoded in an intent graph, and a GNN is trained to recognize the mutual impact of intents.

\section{Conclusions and Future Work}\label{sec:con}

We presented \problemName, an extension to the universal (single intent) \er task, which caters to downstream applications that interpret resolutions in multiple ways. %. The need for multiple resolutions arises when ER is part of a more general data project and multiple intents are involved, representing, 
%{\em e.g.}, when there is a need for resolutions at different granularity levels. 
To tackle this problem, we introduce \modelName, utilizing contemporary solutions to universal \er tasks to solve \problemName. \modelNameSpace trains a graph neural network to learn intents through inter-intent relationships. To show the effectiveness of \modelNameSpace for \problemName, we experimented with \modelNameSpace using three benchmarks, two adapted from existing \er benchmarks along a new benchmark designed for \problemName. We show that \modelNameSpace provides an improvement over the state-of-the-art baseline for the universal \er task.

Future work include extensions of end-to-end \er to support solving \problemName. Specifically, we wish to test \modelNameSpace with additional matchers that produce record pair representations. In addition, we wish to investigate the role of blocking in \problemName.

%We presented \modelname, a novel framework to address. 

%Using four-way expertise characterization, drawing on insights from both matching and metacognition, we provided a novel feature-set to represent a human matcher for the task. We empirically showed the superiority of \emph{MExI} over several state-of-the-art methods. To the best of our knowledge, this work is the first to analyze human expert decision making and mouse movements with LSTMs and CNNs, respectively. Finally, we believe that any human-in-the-loop process may benefit from our framework.
\section{Acknowledgments}\label{sec:ack}
This work was supported in part by the National Science Foundation (NSF) under award numbers IIS- 1956096. We also acknowledge the support of the Benjamin and Florence Free Chair.

%%
%% The next two lines define the bibliography style to be used, and
%% the bibliography file.
%\appendix
%\input{appendix}
%\balance
\bibliographystyle{ACM-Reference-Format}
\bibliography{BIBGranularER.bib}
\balance
\end{document}